\documentclass[journal]{IEEEtran}

\usepackage{multicol}
\usepackage[bookmarks=true]{hyperref}
\usepackage[]{algorithm}
\usepackage{algpseudocode}
\usepackage[absolute,overlay]{textpos} 

\usepackage{amsthm}
\usepackage{mathtools}
\usepackage[thinc]{esdiff}
\usepackage[mathscr]{euscript}

\usepackage{footmisc}
\usepackage{comment}

\DeclareMathOperator*{\argmin}{arg\,min}

\usepackage{balance} 

\usepackage{import}
\usepackage{epsfig} 
\usepackage{times} 
\usepackage{amsmath} 
\usepackage{amssymb}  
\usepackage{bm}
\usepackage{epstopdf}
\usepackage{graphicx}
\usepackage{float}
\usepackage{subcaption}
\usepackage{enumitem}
\usepackage{setspace}
\usepackage[export]{adjustbox}

\usepackage{tikz}
\usetikzlibrary{arrows}

\usepackage{xcolor}
\usepackage[dvips,frame,rotate,import]{xy}

\usepackage{url}

\newtheorem{assumption}{Assumption}
\theoremstyle{definition}
\newtheorem{definition}{Definition}
\newtheorem{problem}{Problem}

\newcommand{\dStar}{D$^*$}
\newcommand{\rrtStar}{RRT$^*$}
\newcommand{\rrtX}{RRT$^X$}
\newcommand{\algName}{PiP-X}

\newcommand{\subroutine}[1]{\texttt{#1}}

\newcommand{\compositeNode}[2]{$\begin{array}{cc} \cline{1-2}
    \multicolumn{1}{|c|}{#1} & \multicolumn{1}{c|}{#2} \\\cline{1-2} \end{array}$}
 
\algdef{SE}[SUBALG]{Indent}{EndIndent}{}{\algorithmicend\ }%
\algtext*{Indent}
\algtext*{EndIndent}

\definecolor{medium-blue}{rgb}{0,0.25,1}
\definecolor{medium-green}{rgb}{0,0.75,0.25}

\newcommand{\figref}[1]{{Fig.~\ref{#1}}}

\newcommand{\eref}[1]{Eq.~\eqref{#1}}
\newcommand{\secref}[1]{Section~\ref{#1}}
\newcommand{\algoref}[1]{Algorithm~\ref{#1}}
\newcommand{\algolineref}[1]{(line~\ref{#1})}
\newcommand{\algomultilineref}[2]{(lines~\ref{#1}$-$\ref{#2})}

\newcommand\tab[1][1cm]{\hspace*{#1}}

\newcommand{\norm}[1]{\left\lVert#1\right\rVert}

\newcommand{\stateVector}{\bm{x}}

\newcommand{\stateSpace}{\mathbb{R}^n}
\newcommand{\funnel}{\mathcal{F}}
\newcommand{\LyapunovFunction}{V(t,\stateVector)}
\newcommand{\LyapunovFunctionDerivative}{\Dot{V}(t,\stateVector)}
\newcommand{\levelSet}{\mathcal{B}}
\newcommand{\ellipsoid}{\mathcal{E}}
\newcommand{\funnelLibrary}{\mathfrak{L}}
\newcommand{\priorityQ}{\mathcal{Q}}

\newcommand{\config}{\mathcal{C}}
\newcommand{\configSpace}{$\mathcal{C}$-space}
\newcommand{\workSpace}{\mathcal{W}}
\newcommand{\obstacleSpace}{\mathcal{O}}

\newcommand{\searchFunnel}{\mathscr{F}}
\newcommand{\searchGraph}{\mathcal{G}}

\newcommand{\vectwointextbody}[2]{\begin{bmatrix} #1 & #2\end{bmatrix}^T}

\begin{document}

   

\title{\algName : Online feedback motion planning/replanning in dynamic environments using invariant funnels}


\author{{Mohamed Khalid~M Jaffar and Michael~Otte}
\thanks{This work is supported by the Naval Air Systems Command (NAVAIR) under the grant N00421-21-1-0001.}
\thanks{The authors are affiliated with the Department of Aerospace Engineering, University of Maryland, College Park, MD 20742, USA. Email: \{\tt\small khalid26, otte\}@umd.edu}}



\maketitle

\begin{abstract}

Computing kinodynamically feasible motion plans and repairing them on-the-fly as the environment changes is a challenging, yet relevant problem in robot-navigation. We propose a novel \emph{online single-query sampling-based motion re-planning} algorithm $-$ \algName, using finite-time invariant sets $-$ \emph{``funnels"}. We combine concepts from sampling-based methods, nonlinear systems analysis and control theory to create a single framework that enables feedback motion re-planning for any general nonlinear dynamical system in dynamic workspaces.


A volumetric \emph{funnel-graph} is constructed using sampling-based methods, and an optimal \emph{funnel-path} from robot configuration to a desired goal region is then determined by computing the shortest-path subtree in it. Analysing and formally quantifying the stability of trajectories using Lyapunov level-set theory ensures kinodynamic feasibility and guaranteed set-invariance of the solution-paths. The use of incremental search techniques and a pre-computed library of motion-primitives ensure that our method can be used for quick online rewiring of controllable motion plans in densely cluttered and dynamic environments.


We represent traversability and sequencibility of trajectories together in the form of an \emph{augmented directed-graph}, helping us leverage discrete graph-based replanning algorithms to efficiently recompute feasible and controllable motion plans that are volumetric in nature. We validate our approach on a simulated 6DOF quadrotor platform in a variety of scenarios within a maze and random forest environment. From repeated experiments, we analyse the performance in terms of algorithm-success and length of traversed-trajectory. \newline 


\end{abstract}


\IEEEpeerreviewmaketitle

\begin{figure}[h!]
    \centering
    \includegraphics[width=\columnwidth]{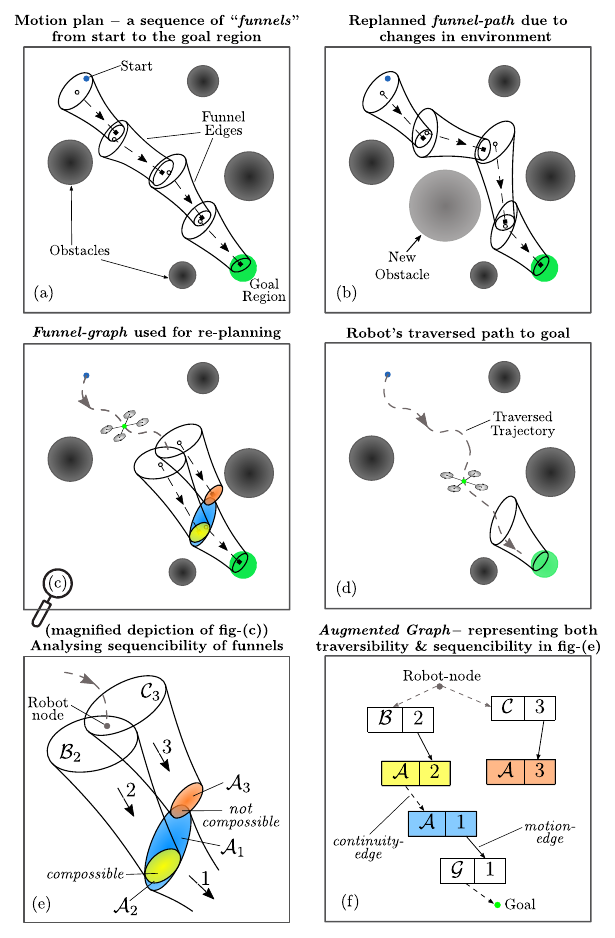}
    \caption{Online funnel-based re-planning algorithm, \algName $-$ (a) Motion plan, a sequence of finite-time invariant sets $-$ \emph{funnels}, each ``dropping" into the subsequent funnel, finally into the goal region. (b) Rewiring of the \emph{funnel-path} when changes in the environment are sensed (c) Underlying search-graph with funnel-edges (d) Trajectory of the robot lies completely inside the traversed funnel-path (e) Analysing sequencibility of funnels (f) Encoding the information of traversability (solid \emph{motion-edges}) and compossibility (dashed \emph{continuity-edges}) in an \emph{augmented graph} to enable quick and efficient graph-based re-planning. The solution path is $\mathcal{B}_2$$-$$\mathcal{A}_2$$-$$\mathcal{A}_1$$-$$\mathcal{G}_1$} 
    \label{fig:visual abstract}
\end{figure}

\section{INTRODUCTION}
\label{sec:introduction}

\IEEEPARstart{T}{he} ability to replan is essential whenever a robot must explore an unknown or changing environment while using a limited sensor radius. In scenarios where the operating workspace has fast-moving obstacles or is densely cluttered with obstacles, a motion-plan must be updated quickly and on-the-fly. Hence, there is a need for fast replanning algorithms. As the obstacle-space changes, the valid state-space also changes, and it is computationally expensive to rebuild the motion plans in such non-convex spaces. Hence, numerical methods such as sampling-based planning (and re-planning) techniques are often used in dynamic and high-dimensional spaces. However, we still require theoretical control techniques to ensure kinodynamic feasibility and trackability of generated motion plans. This dual requirement of motion planning and robot control is considered in tandem by recent literature in \emph{feedback motion planning}.

While brute-force replanning---planning from scratch whenever the environment changes---may work for simple systems, the computational complexity of doing so is often impractical, especially for robots that must react quickly to new information about the changes in the environment. It is much more efficient to reuse the valid portions of previous plans, repairing only the invalid parts to respect the new changes. Incremental search methods which utilise information up to the current iteration to plan/replan in the future can be adapted to achieve quick replanning. This paper presents a novel sampling-based online motion re-planning algorithm using verified trajectories a.k.a. \emph{funnels}; our work utilises a novel graph representation for a  network of funnels that allows the use of incremental graph search for replanning on-the-fly.


Our work extends sampling-based replanners to robust kinodynamic settings and dynamic environments. Using system analysis and invariant set theory, our approach computes dynamically feasible and verified trajectories with formal stability guarantees. The use of sampling techniques enables our algorithm to be computationally tractable for higher dimensional systems and configuration spaces. Additionally, the method is capable of using trajectory libraries to speedup online computation. Hence, our \emph{kinodynamically-feasible}, \emph{online} \emph{re-planning} method finds relevance in practical scenarios such as safely navigating through dynamic environments. The \textbf{novelties} of our algorithm are summarized as follows,
\begin{itemize}
    \item Feedback motion re-planning with funnels, using sampling-based techniques and incremental graph-search
    \item A novel approach to compute motion plans with formal invariance guarantees for any nonlinear non-holonomic robot-system, that can be feasibly tracked by the robot
    \item Representing robot traversability and compossibility of funnels as an \emph{augmented bipartite graph}, enabling the use of discrete graph-search methods to quickly compute kinodynamically-feasible safe motion-plans on-the-fly
    \item Ability to use a pre-computed library of motion primitives with guaranteed regions-of-invariance for quick online re-planning
\end{itemize}

We believe sampling-based planning using funnels is elegant due to its set-invariance, and use of a limited number of volumetric verified-trajectories to probabilistically cover the configuration space. It can also serve as a bridge between discrete numerical methods (path-planning algorithms) and theoretical analysis (system analysis and control design). We can reconcile the two sub-blocks in a robot-autonomy stack---motion planner and controller---using funnels, thus addressing \emph{feedback motion re-planning}. In essence, we get a planning algorithm that respects the closed-loop dynamics of the robot, computes the minimum-cost path to the goal region, and also efficiently replans around dynamic obstacles.

The authors believe this is the first work to propose techniques for \emph{funnel-based motion re-planning} using incremental graph-search. It sets the foundations for feedback motion re-planning on-the-fly by providing a new form of augmented \emph{volumetric} search-graph that is compatible with sampling-based techniques and invariant set analysis. It can additionally provide computation speed-ups using a funnel library. The authors stress that we are not suggesting an improved way to compute funnels, but rather a new technique of using existing efficient methods of funnel computation to motion plan/replan for any general nonlinear robot-system in dynamic spaces.

\subsection{Statement of Contributions}
The \textbf{technical contributions} of our work are three-fold,
\begin{enumerate}
    \item \algName\footnote{PiP-X stands for Planning/replanning in Pipes in dynamic or initially unknown environments} $-$ Online single-query sampling-based \emph{feedback motion re-planning} algorithm using \emph{funnels}.
    \item A novel technique to represent funnel-compossibility using a directed graph embedded within the planning graph, resulting in an \emph{augmented bipartite search-graph}.
    \item Implicitly addressing the \emph{two-point boundary value problem} (TP-BVP) during graph-rewiring by using the system's stability analysis and trajectory sequencibility.
\end{enumerate}

\subsection{Outline}
The remainder of this paper is structured as follows: \secref{sec:related work} outlines the related work, and \secref{sec:preliminaries} introduces the notation while providing the necessary theoretical background. \secref{sec:problem formulation} formally states our problem definition. Our approach is described in \secref{sec:approach} and validated in \secref{sec:validation}. Finally, we state our conclusions in \secref{sec:conclusion}.
\section{RELATED WORK}
\label{sec:related work}

PiP-X builds on existing literature in the fields of feedback motion planning, sampling-based techniques, and online replanning in dynamic environments. It differs from previous work in that it is the first \emph{online} \emph{feedback motion re-planning} algorithm using \emph{funnels}.

\subsection{Sampling-based kinodynamic motion planning}

Geometric sampling-based motion planners such as probabilistic roadmaps (PRM) \cite{Kavraki.etal.PRM96} and rapidly exploring random trees (RRT) \cite{Kuffner.Lavalle.ICRA00} are relevant for planning in high-dimensional dynamical systems. Karaman and Frazzoli \cite{Karaman.Frazzoli.RRT*11} propose RRT$^*$, and provide theoretical proofs of asymptotic optimality. RRT$^\#$ algorithm presented in \cite{Arslan.Tsiotras.RRTSharp13} improves the convergence rate, making it suitable for online implementation. 

In order to address motion planning for differentially constrained robots, numerous researchers have extended such geometric planners to kinodynamic systems \cite{Hsu.etal.IJRR02} \cite{Karaman.Frazzoli.CDC10}. Such algorithms ``steer" the vehicle by randomly sampling control inputs and forward simulating the trajectory based on the dynamics \cite{Lavalle2001Kinodynamic} \cite{Kleinbort.etal.RAL18}. Some researchers formulate an optimal control problem with the trajectory given by the geometric planner, solved using shooting methods \cite{Hwan.Karaman.Frazzoli.CDC11} or closed-form analytical solutions \cite{Webb.Berg.ICRA13}. Another popular approach is to smooth the path given by geometric planners through splines or trajectory optimisation \cite{Ravankar.etal.PathSmoothening} and track it using feedback controllers such as PID or receding-horizon controller \cite{Basescu.Moore.ICRA20}.

\subsection{Motion replanning in dynamic environments}

Earlier work on re-planning $-$ D$^*$ \cite{Stentz95.DStar}, LPA$^*$ \cite{Koenig.etal.AAAI04}, \dStar Lite \cite{Koenig.Likachev.AAAI02} are based on incremental, heuristic-guided shortest-path repairs on a discrete grid embedded in the robot's workspace. Such discretization assumes a constant resolution, requires additional pre-processing, or post-processing to achieve kinodynamic feasibility and/or controllability, and uses data structures that tend to scale exponentially with the dimensions of the system. Nevertheless, they provide a strong algorithmic foundation for developing quick, efficient re-planning algorithms that are useful in cases such as \emph{geometric} path planning, where robot's kinematics, dynamics, or control can be ignored.

Previous work on sampling based replanning focused on the feasibility problem $-$ ERRT \cite{bruce2002.ERRT}, DRRT \cite{Ferguson2006.DRRT} and multipartite RRT \cite{Zucker2007.MultipartiteRRT}, completely pruning the edges in collision and attempting to rejoin the disconnected branches to the rooted tree. \rrtX \cite{Otte.Frazzoli.RRTX16} was the first asymptotically optimal sampling-based \emph{re-planning} algorithm. It rewires the shortest-path subtree from goal to exclude tree nodes and edges that are in collision, similar to \dStar Lite. The underlying search graph---built iteratively through sampling and \emph{rewiring-cascade} step---ensures quick replanning and is well-suited for re-planning on-the-fly. Another technique is to resample configurations based on heuristics \cite{Gammell.etal.IJRR20} and leverage the rewiring step from \rrtStar  to locally repair the solution branch around newly-sensed obstacles \cite{Connell.La.SMC17} \cite{Adiyatov.etal.ICMA17}.

Completeness and optimality guarantees have been achieved for geometric path re-planning, but incorporating robot dynamics without violating these guarantees remains an active area of research. Specifically, most of the optimal sampling-based and re-planning algorithms require solving the two-point boundary-value problem (TPBVP) which is generally difficult for non-holonomic robots---limiting their practical applicability. Some techniques have been proposed to solve without using two-point BVP \cite{Li.Littelfield.Bekris.IJRR16}, achieving near-optimality. However, most of the works consider simple or linearised dynamics without considering disturbances and unmodelled effects. Solving the intrinsic TPBVP for an arbitrary nonlinear system remains challenging and computationally intractable for online implementation.

\subsection{Funnel-based motion planning}

Historically, robot-planning stacks have a hierarchical structure: the high-level path planner computes an open-loop trajectory and a low-level controller stabilises and tracks the trajectory. This decoupled approach is limited in practice because controller tracking errors and actuator saturations or uncertainties might render the planned path infeasible to track. Tracking errors between the planned and actual trajectories can lead to critical failures such as collisions with obstacles. These shortcomings are addressed through \emph{feedback motion planning}, in which the motion planner explicitly considers the stabilising feedback controller to optimise planning for dynamical continuous systems. Mason et al. \cite{mason1985mechanics} introduced a metaphor, ``\emph{funnel}", for locally stabilised and verified trajectories. An illustration of sequentially composed funnels reaching a goal region, similar to \figref{fig:visual abstract}-a, presented in \cite{burridge1999sequential} sparked the motivation to use such funnels for feedback motion planning.

Tedrake et al. \cite{Tedrake.etal.IJRR10} popularised the notion of LQR-trees$-$ an algorithm that covers the state-space using a tree of time-varying trajectories, locally stabilised by an LQR controller and verified by Lyapunov level-set theory. \cite{Tobenkin.etal.IFAC11} presents a detailed approach on how to compute these regions of finite-time invariance using Sum-of-Squares (SoS) programming and bilinear alternations. However, these methods are computationally intensive and are not suitable for scenarios in which the obstacles are not known a priori. Majumdar et al. \cite{Majumdar.Tedrake.IJRR17} compute the funnels offline and use them to plan online for flying a glider through a dense setting. Similar work by the same research group leverages these concepts to develop control algorithms for UAV-perching \cite{Moore.etal.BB14}, double pendulum \cite{Majumdar.Ahmadi.Tedrake.ICRA13}, and a cart-pole \cite{Reist.etal.IJRR16}. Funnel-based motion planning for a robotic arm using adaptive feedback control is presented in \cite{Verginis.etal.arXiv21}. 

\begin{figure}[t]
    \centering
    \includegraphics[width=0.8\columnwidth]{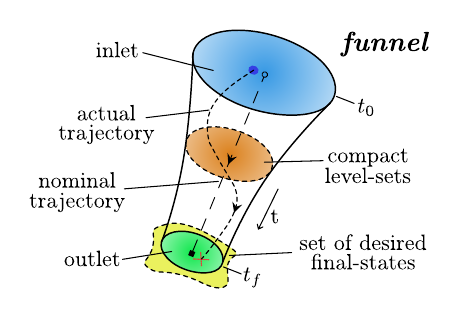}
    \caption{A sample funnel $-$ finite-time backward-reachable invariant set to a compact region of desired final states, $\mathcal{X}_f$}
    \label{fig:sample funnel}
\end{figure}

\subsection{Other related work}
In parallel to such Lyapunov analysis, the controls community has proposed reachability set-based trajectory design \cite{Bajcsy.etal.CDC19} \cite{Kousik.etal.IJRR20}. FaSTrack \cite{Herbert.etal.CDC17} proposes an adversarial game-theoretical approach to generate worst-case tracking error bounds around trajectories using Hamilton-Jacobi reachability analysis. \cite{Singh.etal.IJRR19} uses contraction theory and convex optimisation to compute invariant-tubes around trajectories and plan using them. Other researchers also provide ways to compute stabilised and verified trajectories to be used with any motion-planner - using direct transcription \cite{Manchester.Kuindersma.AuRo19}, and funnel generator functions \cite{Ravanbakhsh.etal.arXiv19}.

Our research extends previous work in \emph{funnel-based motion planning} to \emph{on-the-fly re-planning}. We propose a novel approach to represent funnels and compossibility using graphs, helping us leverage graph-based replanning algorithms to quickly rewire the search tree. Through this, we are able to recast the challenging problem of feedback motion re-planning for non-trivial systems into a geometric one, making it tractable for online planning. Specifically, our work focuses on generating kinodynamically feasible motion plans/replans for higher-order nonlinear systems using backward reachable sets, implicitly solving the two-point boundary value problem.

\section{PRELIMINARIES}
\label{sec:preliminaries}

Our approach combines concepts from invariant set analysis for dynamical systems, design of a library of verified motion primitives, and graph-based replanning techniques. This section briefly provides the requisite background while introducing the notation that would be used in the rest of the paper.

\subsection{Invariant Set Theory}
\label{subsec:invariant set theory}

The notion of region of attraction of asymptotically stable fixed points is extended to certifying time-varying trajectories. Such regions of finite-time invariance around a trajectory are referred to as ``\emph{funnels}" \cite{Tobenkin.etal.IFAC11}. Considering a closed-loop nonlinear system,
\begin{equation}
    \label{eq:closed loop state equation}
    \Dot{\stateVector}(t) = \bm{f}(t,\stateVector(t))
\end{equation}
Where state, $\stateVector \in \stateSpace$ and $\bm{f}$ is Lipschitz continuous in $\stateVector$ and piecewise continuous in $t$. This guarantees global existence and uniqueness of a solution \cite{SlotineBook}. Considering a finite time interval, $[t_0 ,t_f]$, a \emph{funnel} is formally defined as,
\begin{definition}
\textbf{\emph{Funnel}} - A set, $\funnel \subseteq [t_0 ,t_f] \times \stateSpace$, such that for each ($\tau,\bm{x}_\tau$) $\in \funnel$, $\tau \in [t_0 ,t_f]$, the solution to \eref{eq:closed loop state equation}, $\stateVector(t)$, with initial condition $\stateVector(\tau) = x_{\tau}$, lies entirely within $\funnel$ till final time, i.e. $(t,\stateVector(t)) \in \funnel$ $\forall$ $t \in$ $[\tau,t_f]$
\label{def:funnel}
\end{definition}
Intuitively, if the closed-loop system starts within the funnel, then the system states evolving due to \eref{eq:closed loop state equation} remain within the funnel at all time instances until the final time. We leverage tools from Lyapunov theory to compute bounded, inner-approximations of the funnel. The compact level-sets of a Lyapunov function, $\LyapunovFunction$, satisfy the conditions of positive invariance \cite{KhalilBook},
\begin{equation}
    \levelSet(t) = \{\stateVector \mid 0 \leq \LyapunovFunction \leq \rho(t)\}
    \label{eq:Lyapunov level set}
\end{equation}
As noted in \cite{Tobenkin.etal.IFAC11}, under certain mild assumptions, it is sufficient to analyse the boundary of the level sets, $\partial\levelSet(t)$ and the invariance conditions can be reformulated in terms of $\rho(t)$ such as,
\begin{align}
\begin{split}
    &\LyapunovFunction = \rho(t) \implies \LyapunovFunctionDerivative \leq \Dot{\rho}(t) \\
    &\LyapunovFunctionDerivative = \diffp{\LyapunovFunction}{t} + \diffp{\LyapunovFunction}{\stateVector}\bm{f}(t,\stateVector(t))
    \end{split}
    \label{eq:Lyapunov condtion}
\end{align}
In other words, with respect to time, the Lyapunov function at the boundary, $\partial\levelSet(t)$ should decrease faster than the level set, $\rho(t)$ for invariance. Then the set defined as follows, satisfying \eref{eq:Lyapunov condtion} is a \emph{funnel} \cite{Moore.etal.BB14}, 
\begin{equation}
    \funnel = \{(t,\levelSet(t)) \mid t \in [t_0,t_f]\}
    \label{eq:funnel equation}
\end{equation}
In our case, we are interested in computing backward-reachable sets about a finite-time trajectory, to a compact region of state-space. Given a bounded space of desired final-states, $\mathcal{X}_f \subseteq \stateSpace $, we consider funnels that end within the region, i.e. $(t_f,\stateVector(t_f)) \in \funnel \implies \stateVector(t_f) \in \mathcal{X}_f$, or alternatively, $\levelSet({t_f}) \subseteq \mathcal{X}_f$. A sample funnel with its parts labelled can be found in \figref{fig:sample funnel}. 

We wish to maximise the volume of the funnel that flows into the sub-goal region, $\mathcal{X}_f$ using tools from Lyapunov analysis and convex optimisation. The funnel-volume is as defined in \cite{Tobenkin.etal.IFAC11}. \secref{subsec:funnel computation} talks about how to compute such \emph{maximum volume} funnels for systems with piecewise polynomial dynamics, using \emph{Sum-of-Squares} (SoS) relaxation to solve for a class of quadratic Lyapunov functions.

\begin{figure}[t]
    \centering
    \includegraphics[width=0.9\columnwidth]{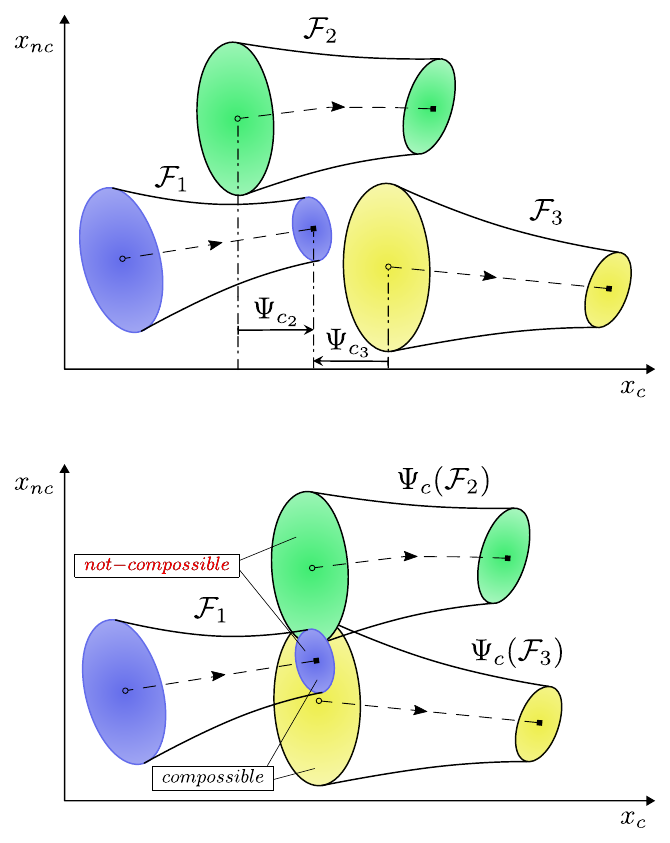}
    \caption{Compossibility of funnels from the library illustrating shifts, $\Psi_c(.)$ along cyclic coordinates (invariant dynamics). ($\mathcal{F}_1$, $\mathcal{F}_3$) is \emph{motion-plan compossible}, whereas ($\mathcal{F}_1$, $\mathcal{F}_2$) is not $-$ i.e. outlet of $\mathcal{F}_1$ is completely contained within the inlet of $\mathcal{F}_3$ after an appropriate shift operation, $\Psi_{c_3}$} 
    \label{fig:composability of funnels}
\end{figure}

\subsection{Verified Trajectory Libraries}
\label{subsec:trajectory library basics}

The idea of saving a pre-computed library of motion primitives for online planning has seen considerable research in the past. An optimisation-based approach to design the library, abstracting the information and sequence of trajectories is provided in \cite{Dey.etal.AAAI12}. Maneuver Automaton \cite{Frazzoli.Dahleh.Feron.TRO05} discusses the relevant properties of trajectory libraries required for sequencing, providing a theoretical foundation. 

The condition for sequencing trajectories presented by \cite{burridge1999sequential} can be extended to funnels by analysing the compact invariant sets satisfying \eref{eq:Lyapunov condtion}, $\levelSet(t)$ at the final and initial time. An ordered pair of funnels, ($\mathcal{F}_i$, $\mathcal{F}_j$) is \emph{sequentially compossible} if $\levelSet_i(t_{f_i}) \subseteq \levelSet_j(t_{0_j})$. However, this is often a strict condition, and is not necessary for composing motion plans. In order to analyse the sequencibility of trajectories for motion planning, we decompose the state vector into cyclic and non-cyclic states, $\stateVector = \vectwointextbody{\stateVector_c^T}{\stateVector_{nc}^T}$. Cyclic states are defined as the coordinates to which the open-loop dynamics of a Lagrangian system, $\Dot{\stateVector} = \bm{f}'(t,\stateVector,\bm{u})$ are invariant, or alternatively, the dynamics depend only on the non-cyclic states, 
\begin{equation}
    \Dot{\stateVector}(t) = \bm{f}'(t,\stateVector_{nc}(t),\bm{u}(t))
    \label{eq:noncyclic states}
\end{equation}

For example, in a 2D disc robot, position is the cyclic coordinates, whereas velocity would be the non-cyclic counterpart. It is sufficient to verify whether the regions-of-invariance projected onto a subspace formed by the non-cyclic state coordinates are sequentially compossible \cite{Majumdar.Tedrake.WAFR13}. One can shift the funnel along the cyclic coordinates so as to contain the outlet of the previous funnel, as illustrated in \figref{fig:composability of funnels}. 

\begin{definition}
A funnel-pair, ($\mathcal{F}_i$, $\mathcal{F}_j$) is said to be \textbf{\emph{motion-plan compossible}} if and only if
\begin{equation}
    \mathcal{P}^\mathcal{S}_{nc}(\levelSet_i(t_{f_i})) \subseteq \mathcal{P}^\mathcal{S}_{nc}(\levelSet_j(t_{0_j}))
    \label{eq:runtime composability}
\end{equation}
\label{def:motion-plan compossibility}
\end{definition}

where $\mathcal{P}^\mathcal{S}_{nc}(.)$ is the projection operator from the state-space onto the subspace formed by non-cyclic coordinates. In addition to posing a less strict condition, the notion of \emph{motion-plan compossibility} plays a significant role in designing the funnel library $-$ one can use a finite number of motion primitives to cover an infinite vector space of cyclic coordinates by shifting the trajectories appropriately. 

For example, during online planning in a UAV, it suffices to check whether the linear and angular velocities at the start of a trajectory match with the current velocities. The position and attitude (cyclic coordinates) of the funnel can be shifted to the current pose of the UAV. The various notions of compossibility is discussed in detail in \cite{Majumdar.Tedrake.IJRR17}, providing methods to check the condition in \eref{eq:runtime composability} using semi-definite programming.

\subsection{Discrete graph replanning}
\label{subsec:discrete graph replanning}
Traditionally, path planners make extensive use of graphs to represent the configuration space $\config$, and deploy various search techniques to search for a feasible or an optimal path through the graph. So, graphs serve a dual purpose of modelling the topology and traversability of $\config$-space, and searching for a solution path through it. 

Graphs, in general, are a mathematical tool to model relations between a group of objects. A graph is defined by a set of vertices and edges, formally denoted as $G = (V,E)$, where $V$ is the set of nodes and set of edges, $E \subseteq \{(v,w) \mid v,w \in V, v \neq w \}$. In typical graph-based and sampling-based motion planning algorithms, vertices and edges represent configurations and trajectories, respectively. In contrast, in this paper, we use graphs to represent a network of \emph{funnels} in a way that respects their volumetric nature and compossibility constraints, \eref{eq:runtime composability}.

Most search algorithms build a connected acyclic subgraph (tree) within the graph, such that each vertex has one, and only one, parent. A path from given start to the goal is readily found by backtracking parent pointers starting from the goal or start depending on the search-direction. Discrete-planners like Dijsktra or A* find an optimal path with respect to a user-defined cost function. The edge costs satisfy the property of a distance metric,

\begin{definition}
\textbf{\emph{Distance metric}} in a set, $\mathcal{V}$, is a function $d_M: \mathcal{V} \times \mathcal{V} \rightarrow \mathbb{R}^+$, to the set of non-negative real numbers. For $v_1,v_2,v_3 \in \mathcal{V}$, the distance metric satisfies the following properties,
\begin{enumerate}
    \item Non-negativity: $d_M(v_1,v_2) > 0$ $\forall$ $v_1 \neq v_2$
    \item Identity of indiscernibles: $d_M(v_1,v_2) = 0$ iff $v_1 = v_2$
    \item Triangle inequality: $d_M(v_1,v_2) + d_M(v_2,v_3) \geq d_M(v_1,v_3)$
\end{enumerate}
\label{def:distance metric}
\end{definition}

Given an edge, $e = (v,w) \in E$, a suitable distance metric $c(v,w)$ represents the cost to move from vertex $v$ to vertex $w$. 
D$^*$ and LPA$^*$ are discrete graph-based \emph{replanning} algorithms that repair the shortest path-to-goal as edge-costs change. Such incremental search techniques reuse all valid current information to improve the solution path in the next iteration, resulting in a faster replanning speed than algorithms that replan from scratch. 

\dStar Lite algorithm continually and efficiently repairs the minimum-cost path from robot to goal, despite changing edge-weights while the robot traverses the path. Our algorithm uses a similar idea to maintain a shortest-path reverse-tree of \emph{funnels} rooted at the goal region.
For each node, the algorithm maintains an estimate of \emph{cost-to-goal} value, $g(v)$, defined as the sum of cost of all edges along the path from node $v$ to the goal, through the graph. In our work, the cost of a \emph{funnel-edge} is given by the length of the nominal trajectory within the funnel. Additionally, the algorithm computes an $lmc$ value (one-step \emph{lookahead minimum cost}) for all nodes, defined as,
\begin{equation}
    lmc(v) = \min_{v' \in N^+} \{c(v,v') + g(v')\}
    \label{eq:lmc definition}
\end{equation}
Where $N^+(v)$ is the set of out-neighbors of vertex, $v$. For $e = (v,w) \in E$, $w$ and $e$ are said to be the out-neighbor and out-edge of $v$, respectively. Similarly, $v$ and $e$ are referred to as the in-neighbor and in-edge of $w$, respectively. 
Based on the two values, $g$ and $lmc$, we determine whether changes have occurred in the shortest path to the goal: $lmc$ is better informed because it gets updated because of changes in out-neighbors' cost-to-goal. The key idea of \dStar Lite can be explained as, 
\begin{enumerate}
    \item $g(v) = lmc(v)$ $\implies$ $v$ is \emph{consistent} $\rightarrow$ no changes to shortest path from $v$ to goal
    \item $g(v) < lmc(v)$ $\implies$ $v$ is \emph{under-consistent} $\rightarrow$ cost has increased, and we have to repair the entire (reverse) subtree rooted at $v$
    \item $g(v) > lmc(v)$ $\implies$ $v$ is \emph{over-consistent} $\rightarrow$ a shorter path exists, update the parent and cost-to-goal of $v$ and propagate this cost-change information to its in-neighbors
\end{enumerate}

In addition to the incremental search, the algorithm uses \emph{heuristics} to focus the reverse-search to the robot-node.
\begin{definition}
\textbf{\emph{Heuristic value}}, $h: \mathcal{V} \rightarrow \mathbb{R}^+$ is a non-negative estimate of the cost from start to a node satisfying
\begin{enumerate}
    \item $h(v) \geq 0$ $\forall$ $v \neq v_{start}$, $h(v_{start}) = 0$
    \item Triangle inequality: $h(v) \leq h(v') + c(v,v')$ $\forall$ $v,v' \in \mathcal{V}$
    \item Admissibility: $h(v) \leq h^*(v)$ $\forall$ $v \in \mathcal{V}$, where \\$h^*(v)$ is the optimal cost-from-start value
\end{enumerate}
\label{def:heuristic}
\end{definition}

We do not repair all the nodes after every edge-cost change, instead only repair \emph{promising nodes} that have the ``potential" to lie in the robot's shortest path to goal, determined using $g$, $lmc$ and $h$ values as in \eref{eq:key for priority queue}. A priority queue is utilised to maintain an order in which nodes need to be repaired. The inconsistent nodes, i.e. $g(v) \neq lmc(v)$, are pushed into the minimum priority queue based on the following key,
\begin{align}
    \begin{split}
      key(v) = [&min\{g(v),lmc(v)\} + h_{start}(v),\\ &min\{g(v),lmc(v)\}]
    \end{split}
\label{eq:key for priority queue} 
\end{align}

In our work, graph-vertices represent regions in state-space$-$ inlet/outlet regions, and edges represent funnels ``flowing" from inlets to outlets. We represent both, traversability and compossibility, using a directed graph and implement incremental search methods to calculate the shortest path from robot to the goal, and recalculate it as the environment changes. This ensures there exists a safe, controllable trajectory starting from the initial state in an \emph{inlet-node} to the final state in an \emph{outlet-node}. 

\section{PROBLEM FORMULATION}
\label{sec:problem formulation}

For a Lagrangian robot system with state, $\bm{x} = (\bm{q},\bm{v}) \in \mathcal{S} \subseteq \mathbb{R}^{2d}$, where $\bm{q} \in \config \subseteq \mathbb{R}^{d}$ is position, $\bm{v} \in \mathbb{R}^{d}$ represents the velocities and $d \in \mathbb{N}$ is the dimension of the configuration space, $\config$, the dynamics are given by,
\begin{equation}
    \dot{\bm{q}} = \bm{v} \tab \dot{\bm{v}} = \bm{f}(\bm{q},\bm{v},\bm{u}) 
    \label{eq:lagrangian robot system}
\end{equation}
where $\bm{u} \in \mathcal{U} \subseteq \mathbb{R}^m$ is the control-input to the robot system, and $\bm{f}$ representing the system dynamics is locally Lipschitz continuous.

\begin{assumption}
For a compact set of desired configurations, $\mathcal{X}_{des} \subseteq \config$, there exists a state-feedback control policy, $\bm{u}: [t_0,t_f] \rightarrow \mathcal{U}$, which when input to the system \eqref{eq:lagrangian robot system}, starting at $\bm{x}(t_0) = \bm{x}_0$, ensures $\bm{q}(t_f) \in \mathcal{X}_{des}$, for some finite time $t_f \geq 0$.

\end{assumption}
We formally quantify the tracking or stabilising performance of this assumed controller by computing inner-approximations of backward-reachable invariant sets around the nominal trajectory, as described in \secref{subsec:funnel computation}.

Consider a robot that operates in workspace, $\workSpace \subseteq \config \subseteq \mathbb{R}^{d}$, with finite number of obstacles having locally Lipschitz continuous boundaries, occupying a subspace, $\obstacleSpace \subset \workSpace$. Correspondingly, let $\config_{obs}$ be the open subset of configurations in which the robot is in-collision. $\config_{free} = \config \setminus \config_{obs}$ is the closed subset of \configSpace, in which the robot can ``safely" operate without colliding with obstacles.

\begin{definition}
\textbf{\emph{Funnel-edge}} - Given a compact set, $\mathcal{X}_w \subseteq \config$ centered around $w \in \config$, an initial configuration, $v \in \config$ and finite time interval, $[t_0, t_f]$ \emph{funnel-edge} $\phi(v,w) \subseteq \config$ is the projection of \emph{maximum-volume} \hyperref[def:funnel]{funnel}, $\funnel$ satisfying \eqref{eq:Lyapunov level set}-\eqref{eq:funnel equation}, such that $v \in \mathcal{P}^\mathcal{S}_\config(\mathcal{B}(t_0))$ and  $\mathcal{P}^\mathcal{S}_\config(\mathcal{B}(t_f)) \subseteq \mathcal{X}_w$. A funnel-edge is said to be \textbf{\emph{valid}}, if and only if $\mathcal{P}^\config_\workSpace(\phi(v,w)) \cap \obstacleSpace = \emptyset$. The cost of the funnel-edge is given by the length of the nominal trajectory, $\bm{x}_0(t)$ projected down to $\config$-space, $\bm{q}_0(t)$
\begin{equation}
     c_\phi(v,w) = \int_{t_0}^{t_f} ds(t)
\label{eq:funnel-edge cost}
\end{equation}
\end{definition}

where $ds = \norm{\bm{dq_0}}$, $\bm{q_0}$ satisfies \eqref{eq:lagrangian robot system}, and $\norm{\cdot}$ represents the Euclidean norm. $\mathcal{P}^A_B(.)$ is the projection operator from a space $A$ to a lower dimensional subspace, $B$. 


It is worth mentioning that the defined cost function \eqref{eq:funnel-edge cost} satisfies the properties in \eqref{def:distance metric}, and hence is a distance metric. Since a valid funnel-edge does not intersect with the obstacle-set, the trajectory contained within it due to set-invariance, will not be in collision with obstacles. Thus, $\bm{q}(t) \in \phi_{valid}(.) \Leftrightarrow \bm{q}(t) \in \config_{free}$ for $t_0 \leq t \leq t_f$.


\begin{definition}
\textbf{\emph{Funnel path}} - For a configuration, $q_1 \in \config_{free}$ and a compact set, $\mathcal{X}_2 \subseteq \config_{free}$, \emph{funnel-path}, $\pi(q_1,\mathcal{X}_2)$ is a finite sequence of \emph{valid} funnel-edges with underlying \hyperref[def:motion-plan compossibility]{\emph{motion-plan compossibility}}, i.e. $\pi(q_1,\mathcal{X}_2) = \{\phi_1, \phi_2, \dots \phi_n\}$, such that $q_1 \in \mathcal{P}^\mathcal{S}_\config(\mathcal{B}_{\phi_1}(t_{0_1}))$ and $\mathcal{P}^\mathcal{S}_\config(\mathcal{B}_{\phi_n}(t_{f_n})) \subseteq \mathcal{X}_2$. The cost of the funnel path is defined,
\begin{equation}
    c_\pi(q_1,\mathcal{X}_2) = \sum_{i=1}^{n} c_{\phi_i}
\end{equation}
\label{def:funnel path}
\end{definition}

\begin{problem}
\textbf{Online motion planning} - Given $\config_{free}$, obstacle space, $\obstacleSpace(t)$, and a goal region, $\mathcal{X}_{goal} \subseteq \config_{free}$ for a robot starting at a configuration, $q_{robot}(0) = q_{start} \in \config_{free}$, calculate the \emph{optimal funnel path}, $\pi^*(q_{robot}(t),\mathcal{X}_{goal})$, move the robot by applying a feedback control policy, $u:(\bm{x},\bm{x}_0,t) \rightarrow \mathcal{U}$ and keep updating $\pi^*$ until $q_{robot}(t) \in \mathcal{X}_{goal}$
\begin{equation*}
    \pi^*(q_{robot}(t),\mathcal{X}_{goal}) = \argmin_{\pi(q_{robot}(t),\mathcal{X}_{goal})} c_\pi(q_{robot}(t),\mathcal{X}_{goal})
\end{equation*}
\end{problem}

A dynamic environment has its obstacle set changing randomly with time and/or with robot position, $\Delta\obstacleSpace(t) = s(t,q_{robot}(t))$. An environment can be modelled as static if $h$ is known \emph{a priori} or can be deterministically computed. A trivial case of static environment is $\Delta\obstacleSpace(t) \equiv \emptyset$. We consider replanning in dynamic environments, where $\Delta\obstacleSpace(t)$ is neither known \emph{a priori} nor possible to predict.
\begin{problem}
\textbf{Motion replanning} - Assuming that a robot has the ability to sense obstacle changes, $\Delta\obstacleSpace(t)$, continually recompute $\pi^*(q_{robot}(t),\mathcal{X}_{goal})$ until $q_{robot}(t) \in \mathcal{X}_{goal}$.
\end{problem}

We propose techniques and algorithms to plan/replan \emph{minimum-cost funnel-paths} on-the-fly in real-time. The funnel-path is a sequence of maneuvers with formal guarantees of invariance, associated with state-feedback tracking control policies. The motion plan/replan with invariant sets and control policies ensures that the robot trajectory lies within the funnel path, avoiding dynamic obstacles and ultimately reaching the goal region.

\begin{figure*}[t]
    \centering
    \includegraphics[width=0.83\textwidth]{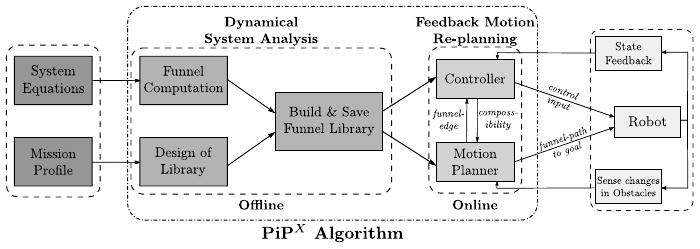}
    \caption{Overview of PiP-X algorithm: consists of an offline stage of dynamical system analysis, and an online phase of sampling-based graph construction and incremental re-planning in dynamic environments}
    \label{fig:algorithm overview}
\end{figure*}

\section{APPROACH}
\label{sec:approach}
This section details the various components of our method: the pre-processing stage of computing backward-reachable invariant sets and designing the funnel library, and the online phase of the motion re-planning algorithm, \algName.

\subsection{Outline}
\label{subsec:high-level overview}

This subsection contains a brief high-level outline of our entire approach, which involves: invariant set analysis, designing a funnel library, and the online feedback motion re-planning using an augmented directed graph. Each part is discussed in greater detail in its own subsection. As illustrated in \figref{fig:algorithm overview}, our method has an offline stage of nonlinear system analysis using Lyapunov theory, and an online phase of sampling-based motion re-planning using incremental graph-search. Given the mission profile, we design the funnel library as described in \secref{subsec:funnel library design}. Design considerations include description of cyclic/non-cyclic coordinates of the state-space for \emph{funnel compossibility}, $\config$-space topology to ensure \emph{probabilistic coverage}, and desired resolution of motion-plan.

We compute backward-reachable invariant sets, detailed in \secref{subsec:funnel computation}, using Lyapunov theory. The system's equations of motion along with the state-feedback control law are approximated to polynomial dynamics about the nominal trajectory using Taylor-series expansion of order greater than 1. We find that such a bounded polynomial approximation offers a conservative estimate, i.e. it always under-estimates the inner-approximations of backward-reachable sets---a condition sufficient for our re-planning algorithm. Given an initial state, a compact set of desired final-states and a finite-time horizon, we calculate the certified region of invariance characterised by Lyapunov level-sets centered around the nominal-trajectory. Due to the computational complexity of nonlinear system analysis, we \emph{pre-construct} a library of verified trajectories with various combinations of initial and final states, to be used during the online-phase of feedback motion re-planning.

A graph representing a network (or roadmap) of \emph{funnel-connectivity} is incrementally built using sampling methods (RRG), and motion plans/replans are computed through it. The graph embedded in the $\config$-space is constructed based on the aforementioned system analysis in the higher dimensional state-space. The graph-edges represent funnels, and vertices represent inlet or outlet regions in state-space, both projected down to the $\config$-space. We analyse the \emph{motion-plan compossibility} (Definition \ref{def:motion-plan compossibility}) of funnel-pairs and additionally include that information in the form of an \emph{augmented search graph} (see \figref{fig:composite graph} and \secref{subsec:augmented graph}). With this graph we compute a shortest-path subtree of funnels, rooted at the goal region using an approach similar to one introduced in \dStar Lite \cite{Koenig.Likachev.AAAI02}. 

Using graph-based replanning methods, motion plans are quickly recomputed in the event of changes in obstacle-space, $\Delta \obstacleSpace$, either due to robot sensing new obstacles or the obstacles being dynamic themselves. The funnel-path to goal region and the corresponding sequence of control-inputs are input to the robot, with state-observer and obstacle-sensor closing the feedback-loop. \secref{subsec:planning algorithm} describes our feedback motion planning/replanning algorithm in-depth.


From \figref{fig:invariance}, it is worth noticing that the funnels computed based on Lyapunov theory offer a sufficient but not a necessary condition for invariance. Trajectories starting inside the funnel will remain in the funnel for the entire finite time-horizon. However, trajectories starting outside the funnel may or may not terminate within the defined goal region. Nevertheless, this analysis provides formal guarantees about robustness to set of initial conditions and system perturbations, pertinent in sampling-based motion planning/replanning of kinodynamic systems.

\subsection{Notes on augmented search-graph}
\label{subsec:augmented graph}

\begin{figure*}[t]
    \centering
    \includegraphics[width=0.8\textwidth]{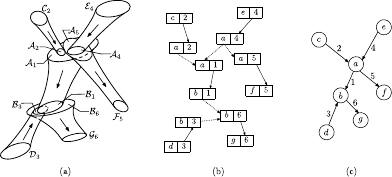}
    \caption{(a) Search funnels existing in the $\mathbb{R}^+ \times \stateSpace$ (b) \emph{Augmented search graph} representing planning in $\config$-space with compossibility information. Solid edges are finite-cost \emph{motion-edges} representing traversability, dashed-edges are zero-cost \emph{continuity-edges} encoding compossibility information $-$ whether trajectories ``flow" into the subsequent funnel (c) Conventional search graph used by typical tree/graph-based motion-planners}
    \label{fig:composite graph}
\end{figure*}

We represent the network of funnels using a directed graph \emph{augmented} with the additional information of funnel-compossibility (see \figref{fig:composite graph}). Such an augmented graph representation enables the use of incremental graph-replanning techniques to quickly rewire funnel-paths (Definition \ref{def:funnel path}) to the goal region. Our method essentially constructs ``links" between regions of state-space with funnels that have an implicit notion of time. Traversability of the robot system and sequencibility of trajectories is represented through \emph{motion-edges} and \emph{continuity edges}, respectively in the \emph{augmented graph} $\searchGraph$. The edge-set of this graph consists of motion-edges and continuity-edges, $E = E_m \cup E_c$. The graph-vertex is a tuple consisting of the configuration and the funnel-edge, $v$ $-$ \compositeNode{q}{f}. The \emph{composite nodes} exhibit certain relations amongst the node-set $V$, summarised as,

\begin{itemize}
    \item \compositeNode{q}{f_1} \& \compositeNode{q}{f_2} $-$ Sharing the same first element (configuration) implies there will be a directed zero-cost \emph{continuity edge}, $e_c$ between the two composite nodes, if and only if ($f_1$, $f_2$) is \hyperref[def:motion-plan compossibility]{\emph{motion-plan compossible}}.
    \item \compositeNode{q_1}{f} \& \compositeNode{q_2}{f} $-$ Represents the case of two vertices sharing a funnel-edge, $f$. Hence, by default a \emph{motion-edge}, $e_m$ exists, having a cost as defined in \eref{eq:funnel-edge cost}
    \item \compositeNode{q_1}{f_1} \& \compositeNode{q_2}{f_2} $-$ These nodes do not have a common entry. In such cases, there will be \emph{no} edge between the 2 composite-nodes. 
\end{itemize}

It is worth mentioning that (dashed) continuity-edges are zero-cost and have no bearing on the cost of the solution path to goal or the optimality guaranteed by the graph-search algorithm. As illustrated in \figref{fig:composite graph}-c, when nodes with the same configuration (first element of the tuple) are grouped together into one \emph{super-node} result in a search-graph with just (solid) motion-edges, usually used by typical path-planners.

As observed in \figref{fig:composite graph}-b, there are two types of graph vertices $-$ \emph{inlet-nodes} and \emph{outlet-nodes}, $V = \{V_I,V_O\}$. Inlet-nodes like \compositeNode{a}{1} have one, and only one, solid out-edge $-$ might have one or more dashed in-edges. Outlet nodes have one, and only one, solid in-edge, and one or more dashed out-edges, e.g. \compositeNode{a}{4}. For any node, the list of in-edges will have either dashed edges or one solid edge. The same is true for out-edges. 

With these properties and observations, the authors would like to point out that the \emph{augmented graph} is indeed \emph{bipartite} $-$ disjoint sets of inlet-nodes and outlet-nodes. Motion-edges go from $V_I$ to $V_O$, and continuity-edges from $V_O$ to $V_I$. Hence, any path from a configuration to the goal will have alternating (solid) motion-edges and (dashed) continuity-edges. 


\subsection{Online motion planning-replanning algorithm $-$ \algName}
\label{subsec:planning algorithm}


A reverse-search graph is more effective in scenarios that require online replanning, such as a robot navigating through an unknown environment perceiving obstacles within a limited sensor radius. It is efficient because it suffices to alter the motion plans only locally near the robot location, saving us the cost of rewiring the bulk of the search-tree. We \hyperref[subsec:constructRRG]{incrementally build the \emph{funnel-graph}} and \hyperref[subsec:computeShortestPath]{compute the optimal \emph{funnel-path}} through the graph using the routine $-$ \subroutine{plan()} (\algoref{algo:planning}). The pre-planning process on a higher level is as follows, 
\begin{enumerate}
    \item Sample a configuration $q_{rand}$ \algolineref{algoline:sample} and extend an $\epsilon$-distance \algolineref{algoline:extend} from the nearest node in the existing search graph to determine new configuration $q_{new}$
    \item Determine the set of nearest neighbors \algolineref{algoline:nearest neighbors} in the shrinking $r$-ball \cite{Karaman.Frazzoli.RRT*11}, $r = \min\{r_0(log|V|/|V|)^{1/d},\epsilon\}$, where $|V|$ is cardinality of the vertex set, $d$ is the dimensionality of $\config$ and $r_0$ is a user-specified parameter
    \item From the funnel library $\funnelLibrary$, choose the trajectory that would steer (\algoref{algo:steering}) the robot from $n$ to a $\delta$-ball near $q_{new}$ as well as the return trajectory from $q_{new}$ to a $\delta$-ball near $n$, for all $n$ in the set of nearest neighbors
    \item Once we get the \emph{funnel-edges}, $\phi$, check for ``overlap" with obstacles, and compossibility among the funnels. Represent the projections of inlets, $\mathcal{X}_i$ and outlets, $\mathcal{X}_o$ as nodes, $\phi$ as a \emph{motion-edge}, and zero-cost \emph{continuity-edges} signifying compossibility $-$ hence is the search-graph iteratively constructed (\algoref{algo:constructRRG})
    \item An incremental search (\algoref{algo:computeShortestPath}) on the constructed sampling-based graph keeps updating the shortest-path subtree rooted at the goal
\end{enumerate}

\begin{algorithm}
 \caption{$(\searchFunnel, \searchGraph) \gets \subroutine{plan}()$}
 \begin{algorithmic}[1]
    \State $q_{rand} \gets \subroutine{sampleFree}()$ \label{algoline:sample} 
    \State $q_{new} \gets \subroutine{extend}(\searchGraph,q_{rand}, \epsilon$) \Comment{geodesic-distance} \label{algoline:extend}
    
    \State $r \gets \subroutine{rBall}()$
    \State $\mathcal{N} \gets \subroutine{findNearestNeighbors}(q_{new},r,\searchGraph)$ \label{algoline:nearest neighbors}
    \State $(\searchFunnel, \searchGraph) \gets \subroutine{constructSearchGraph}(q_{new},\mathcal{N})$ \label{algoline:constructGraph}
    
    \State $\subroutine{computeShortestPathTree}()$ \label{algoline:computeInitialPath}
    \State \textbf{return} $(\searchFunnel, \searchGraph)$
 \end{algorithmic} 
 \label{algo:planning}
\end{algorithm}

\begin{algorithm}
 \caption{\algName}
 \begin{algorithmic}[1]
    \Require $q_{start}$, $\mathcal{X}_{goal}$, $\config_{free}$,  $\obstacleSpace$, $\funnelLibrary$ \Comment{Start, Goal region, Free- \tab[2.6cm] space, obstacle-space, Funnel library}
    \Ensure $\searchGraph, \searchFunnel$  \Comment{search-Graph and search-Funnel} 
    \\ \textbf{\emph{Parameters}}: $\epsilon$, $r_0$, $T_P$, $I_M$ \label{algoline:initialise} \Comment{extend distance, $r$-ball, \tab[3.5cm] pre-planning time, \emph{idleness} limit}
    \\ \textbf{\emph{Initialisation}}: $t \gets 0$, $startFound \gets 0$, \\ \tab[2cm] $\searchGraph$.\subroutine{add}($\mathcal{X}_{goal}$), $\searchFunnel$ $\gets$ $\emptyset$ \label{algoline:end initialise}
    
    \While{$t < T_P \lor \neg startFound$} \label{algoline:pre-planning}
        \State $(\searchFunnel, \searchGraph) \gets \subroutine{plan}()$
        \If{$\subroutine{inFunnel}(q_{start},\searchFunnel)$} \label{algoline:inFunnel check} 
            \State $startFound \gets 1$
        \EndIf
    \EndWhile \Comment{end of \textbf{Pre-planning} phase} \label{algoline:end pre-planning}
    \State $j \gets 0$, $q_{robot} \gets q_{start}$, $q_{prev} \gets q_{start}$ %
    \While{$j < I_M \land q_{robot} \notin \mathcal{X}_{goal}$} \Comment{\textbf{Online} phase}
        
    \State \textbf{at} \emph{sensingFrequency} \textbf{do} \Comment{Sensing obstacle-changes} \label{algoline:sensing start}
            \State \tab[0.4cm] $\Delta \obstacleSpace \gets \subroutine{senseObstacles}()$
            \State \tab[0.4cm] \subroutine{modifyEdgeCosts($\Delta \obstacleSpace$)}
    \State \textbf{end} \label{algoline:sensing end}
        
        \State $(\searchFunnel, \searchGraph) \gets \subroutine{plan}()$ \Comment{\emph{Repairing} the motion-plan} \label{algoline:replanning}
        
        \State \textbf{at} \emph{robotMotionFrequency} \textbf{do} \Comment{Robot movement} \label{algoline:motion start}
        \Indent \vspace{-1mm}
            \State $q_{robot} \gets \subroutine{robotMove}(q_{robot},q_{goal})$ \label{algoline:robot move}
            \If{$g(q_{robot}) \neq \infty$} \Comment{a \emph{funnel-path} exists}
                \State $k_m \gets k_m + \subroutine{computeHeuristic}(q_{prev},q_{robot})$
            \State $q_{prev} \gets q_{robot}$; $j \gets 0$ \Comment{reset \emph{idleness} count}
             \Else 
             \State $q_{robot} \gets q_{prev}$ \Comment{stay at current location}
             \State $j \gets j+1$ \Comment{update \emph{idleness} count} 
            \EndIf
        \EndIndent \vspace{-1mm}
        \State \textbf{end} \label{algoline:motion end}
        
        
        \If{$q_{robot} \in \mathcal{X}_{goal}$}
            \State \textbf{return} $\texttt{SUCCESS}$ \Comment{Algorithm success}
        \EndIf
    \EndWhile
    
    \State \textbf{return} $\texttt{NULL}$ \Comment{Algorithm failure}
    
 \end{algorithmic}
 \label{algo:main algorithm}
\end{algorithm}

After each update \algolineref{algoline:computeInitialPath}, all the consistent nodes in the graph know their best parent according to \eref{eq:determine parent node}, enabling the planner to backtrack the solution-path using parent pointers. The search is focused towards the robot location using an admissible heuristic, $h(v)$ as defined in \eqref{def:heuristic}, thereby enabling quick rewiring of the optimal path whenever the heuristic provides useful information.

The pre-planning phase in \algoref{algo:main algorithm} (lines \ref{algoline:pre-planning}$-$\ref{algoline:end pre-planning}) continues until the start configuration lies within one of the funnel-inlets and the search-graph is dense enough to have covered a sufficient volume of the \configSpace. A solution funnel path exists if the robot configuration lies within one of the inlet-nodes and has a finite cost-to-goal value. Consequently, we have a sequence of closed-loop control policies to transition from a region of state-space to another, ultimately terminating at the goal region. The entire trajectory is guaranteed to lie within the solution funnel branch by the virtue of set-invariance, provided the actual model sufficiently resembles the nominal model.

The various routines of our \emph{online re-planner}, PiP-X (\algoref{algo:main algorithm}) providing low-level implementation details, are explained as follows: We first specify the algorithm parameters and inputs, and initialise the required data structures $-$ graph, kdTree, priority queue (lines \ref{algoline:initialise}$-$\ref{algoline:end initialise}). The planning parameters are minimum path-resolution, $\epsilon$, shrinking $r$-ball parameters, $r_0$ and $d$, pre-planning time, $T_P$ and idleness limit, $I_M$. The inputs to the algorithm are start configuration, $q_{start}$, goal region, $\mathcal{X}_{goal}$, the pre-computed funnel library, $\funnelLibrary$, and the initial environment $-$ characterised through $\config_{free}$ and list of obstacles known \emph{a priori}, $\obstacleSpace$. The obstacle-space will be updated when any changes, $\Delta \obstacleSpace(t)$ are discovered on-the-fly. 

\subsubsection{\textbf{Sampling configurations}} 
\label{subsec:sampling}
\subroutine{sampleFree}() $-$ The configurations $q_{rand}$ are independent and identically (i.i.d.) drawn from the free-space, $\config_{free}$ at random. When the robot starts moving and senses obstacles, the sampling is directed towards the sensed region where changes are certain to have occurred. This helps to rewire the parent-edges near the robot, ensuring the robot has a choice of safe alternate plans around the new-found obstacles. However, configurations are continued to be drawn uniformly random from $\config_{free}$ at regular frequency even after the robot starts moving, for probabilistic coverage and completeness.  

\subsubsection{\textbf{Graph extension}}
The $\config$-space is explored using the \subroutine{extend}($\searchGraph$, $q_{rand}$, $\epsilon$) routine. It determines the nearest configuration in the existing search graph, based on geodesic distance, and aims to extend to $q_{rand}$ by at most an $\epsilon$-distance to obtain the new configuration, $q_{new}$. If this configuration is already in the search-funnel, $\searchFunnel$, we discard it and continue with the next sampling, because we are guaranteed to find a set of maneuvers which would drive the robot-system from this configuration to the goal region. 

\subroutine{findNearestNeighbors}($q_{new}$, $r$, $\searchGraph$) determines the neighbors within a $r$-ball around the new configuration, $q_{new}$. It is implemented through a k-D tree built using configurations. The radius of the ball decreases at a ``shrinking rate" derived using percolation theory \cite{Penrose.RRG.2003}.

\begin{algorithm}
 \caption{$\funnel \gets \subroutine{steer}(q_1,q_2)$}
 \begin{algorithmic}[1]
    \State $\funnel' \gets \subroutine{findFunnel}(q_1,q_2,\funnelLibrary)$
    \State $\funnel \gets \subroutine{shiftFunnel}(q_2,\funnel')$ \Comment{shifts \& truncates the funnel}
    \If{$q_1 \notin \funnel.ellipsoid(\subroutine{start}) \lor \neg \subroutine{collisionFree}(\funnel,\obstacleSpace)$} \label{algoline:funnel-collision check}
        \State \textbf{return} $\emptyset$
    \EndIf
    \State \textbf{return} $\funnel$ 
 \end{algorithmic} 
 \label{algo:steering}
\end{algorithm}

\subsubsection{\textbf{Steering} (\algoref{algo:steering})}
\label{subsec:steering}
From the motion-primitives library, \subroutine{findFunnel}() determines the appropriate funnel that closely drives the system from configuration, $q_1$ to $q_2$. We use \subroutine{shiftFunnel}() subroutine to shift the funnel along the cyclic coordinates and time, using appropriate shift-operators $\Psi_c(.)$ and $\Psi_t(.)$, respectively. The fact that the funnel is a backward-reachable set, enables us to truncate the funnel at any time, $t_f \in [0, T)$. 

The funnel is projected down to the workspace for checking any overlaps with the obstacle-set, $\obstacleSpace$. If the funnel is \hyperref[subsec:collision checking]{\emph{in-collision}} or the target-configuration does not lie in the inlet of the funnel projected down to $\config$ \algolineref{algoline:funnel-collision check}, the subroutine \subroutine{steer}() returns a null set. Otherwise, we return the funnel, $\funnel$ along with its cost, computed as the length of the nominal trajectory within the funnel according to \eref{eq:funnel-edge cost}.

\begin{algorithm}
 \caption{$(\searchFunnel, \searchGraph) \gets \subroutine{constructSearchGraph}(q_{new}, \mathcal{N}$)}
 \begin{algorithmic}[1]
    \ForAll{$n \in \mathcal{N}$}
        
        \State  $\funnel^-_n \gets \subroutine{steer}(q_{new},n)$ \Comment{funnels out of $q_{new}$} \label{algoline:graph-steer1}
        \State  $\funnel^+_n \gets \subroutine{steer}(n,q_{new})$ \Comment{funnels into $q_{new}$} \label{algoline:graph-steer2}
        
        \If {$\funnel^-_n \neq \emptyset$}
            \State $\{\mathcal{X}_i,\mathcal{X}_o\} \gets \subroutine{getNode}(\funnel^-_n)$ \Comment{inlet-outlet node} \label{algoline:composite node1}
            
            \ForAll{$\funnel_o \in outFunnels(n) \cup \{\funnel^+_n\}$} \label{algoline:outFunnel compossibility start}
            \If{$\subroutine{compossible}(\funnel^-_n,\funnel_o)$}
                \State $\mathcal{N}_i \gets inletNode(\funnel_o)$ 
                \State $E_c \gets E_c \cup {(\mathcal{X}_o,\mathcal{N}_i)}$ \Comment{continuity-edge} \label{algoline:add continuity-edges 1}
                \EndIf
            \EndFor \label{algoline:outFunnel compossibility end}
            
            \State $V \gets V \cup \{\mathcal{X}_i,\mathcal{X}_o\}$; $E_m \gets E_m \cup (\mathcal{X}_i,\mathcal{X}_o)$ \label{algoline:add vertices and motion-edges 1}
            \State $\subroutine{updateVertex}(\mathcal{X}_o)$ \label{algoline:propagation}
        \EndIf
        
        \If {$\funnel^+_n \neq \emptyset$}
            \State $\{\mathcal{X}_i,\mathcal{X}_o\} \gets \subroutine{getNode}(\funnel^+_n)$ \Comment{inlet-outlet node} \label{algoline:composite node2}
            
            \ForAll{$\funnel_i \in inFunnels(n) \cup \{\funnel^-_n\}$} \label{algoline:inFunnel compossibility start}
            \If{$\subroutine{compossible}(\funnel_i,\funnel^+_n)$}
                \State $\mathcal{N}_o \gets outletNode(\funnel_i)$ 
                \State $E_c \gets E_c \cup {(\mathcal{N}_o,\mathcal{X}_i)}$ \Comment{continuity-edge} \label{algoline:add continuity-edges 2}
                \EndIf
            \EndFor \label{algoline:inFunnel compossibility end}
            
            \State $V \gets V \cup \{\mathcal{X}_i,\mathcal{X}_o\}$; $E_m \gets E_m \cup (\mathcal{X}_i,\mathcal{X}_o)$ \label{algoline:add vertices and motion-edges 2}
        \EndIf
        \State $\searchFunnel \gets \searchFunnel \cup \{\funnel^-_n, \funnel^+_n\}$ \Comment{adding to funnel-edges set} \label{algoline:update funnel-set}
    \EndFor
    
    \ForAll{$\funnel_i \in inFunnels(q_{new})$} \label{algoline:existing funnels compossibility start}
        \ForAll{$\funnel_o \in outFunnels(q_{new})$}
            \If{$\subroutine{compossible}(\funnel_i,\funnel_o)$}
                \State $\mathcal{X}_o \gets outletNode(\funnel_i)$ 
                \State $\mathcal{X}_i \gets inletNode(\funnel_o)$ 
                \State $E_c \gets E_c \cup {(\mathcal{X}_o,\mathcal{X}_i)}$ \Comment{continuity-edge} \label{algoline:add continuity-edges 3}
            \EndIf
        \EndFor
    \EndFor \label{algoline:existing funnels compossibility end}
    \State \textbf{return} $\searchFunnel$, $\searchGraph = (V, E_m, E_c)$ 
  \end{algorithmic}
  \label{algo:constructRRG}
\end{algorithm}

\subsubsection{\textbf{Constructing the search funnel-graph} (\algoref{algo:constructRRG})}
\label{subsec:constructRRG}
We attempt to construct funnels \algomultilineref{algoline:graph-steer1}{algoline:graph-steer2} between the new configuration, $q_{new}$ and all of the nodes in the neighbor-set, $n \in \mathcal{N}$. The valid funnels flowing into a $\delta_o$-ball around $q_{new}$ are referred to as its $inFunnels$, $\funnel^-_q \equiv \funnel^+_n$ and funnels flowing out of a $\delta_i$-ball around $q_{new}$ as its $outFunnels$, $\funnel^+_q \equiv \funnel^-_n$, $\forall$ $n \in \mathcal{N}$. 

\subroutine{getNode}($\funnel$) in lines \ref{algoline:composite node1} and \ref{algoline:composite node2} determines the nodes, $\mathcal{X}_i$ and $\mathcal{X}_o$, corresponding to the inlet and outlet of the funnel projected down to \configSpace. These nodes are added to the set of graph-vertices, $V$ and the directed edge, ($\mathcal{X}_i$, $\mathcal{X}_o$), is added to the set of \emph{motion-edges}, $E_m$ (lines \ref{algoline:add vertices and motion-edges 1} and \ref{algoline:add vertices and motion-edges 2}) The new constructed funnels are added to the search-funnel, $\searchFunnel$ \algolineref{algoline:update funnel-set}. 

All the pairs of inFunnels and outFunnels at $q_{new}$ are checked for sequencibility (lines \ref{algoline:outFunnel compossibility start}$-$\ref{algoline:outFunnel compossibility end} and lines \ref{algoline:inFunnel compossibility start}$-$\ref{algoline:inFunnel compossibility end}) using \subroutine{compossible}() (\algoref{algo:compossibilty check}). If compossible, a zero-cost directed edge from outlet-node to inlet-node is added to the set of \emph{continuity-edges}, $E_c$ (lines \ref{algoline:add continuity-edges 1} and \ref{algoline:add continuity-edges 2}). Additionally, the existing inFunnels and outFunnels at neighbor-nodes, $n$ are checked for compossibility with the new constructed funnels to/from $n$ \algomultilineref{algoline:existing funnels compossibility start}{algoline:existing funnels compossibility end}. If compossible, zero-cost continuity-edges between outlet-nodes and inlet-nodes at $n$ are added to $E_c$ \algolineref{algoline:add continuity-edges 3}.

Invoking \subroutine{updateVertex}() (\algoref{algo:update vertex}) in line \ref{algoline:propagation} ensures propagation of cost-changes and possible rewiring of the shortest-path subtree due to the new sample. The cost-to-goal value of all the new nodes, $g(v)$ is initialised to be infinite by default. By the virtue of the nodes being inconsistent (specifically overconsistent), they are pushed into the priority queue, $\priorityQ$ with key computed as in \eref{eq:key for priority queue}, and will be repaired if they have the ``potential" to lie in the solution path to goal.

\begin{algorithm}
 \caption{\subroutine{compossible}($\funnel_1,\funnel_2$)}
 \begin{algorithmic}[1]
    \State $\ellipsoid_i \gets \funnel_2(\subroutine{start}).ellipsoid$ \Comment{inlet of $\funnel_2$}
    \State $\ellipsoid_o \gets \funnel_1(\subroutine{end}).ellipsoid$  \Comment{outlet of $\funnel_1$}
    \If{$\ellipsoid_i$ $\supseteq$ $\ellipsoid_o$} \Comment{inlet completely contains outlet} \label{algoline:ellipsoid-in-ellipsoid check}
        \State \textbf{return} \subroutine{TRUE}
    \EndIf
    \State \textbf{return} \subroutine{FALSE}   
  \end{algorithmic} 
  \label{algo:compossibilty check}
\end{algorithm}

\subsubsection{\textbf{Funnel-related subroutines}}
\label{subsec:funnel-related subroutines}
The re-planning algorithm makes use of minor subroutines specific to \emph{funnels}. \subroutine{inFunnel}($q$, $\searchFunnel$) returns a boolean value, based on whether the configuration $q$ lies in any of the inlets of the funnel-edges in $\searchFunnel$. This is useful while checking whether a path exists from start to goal region (\algoref{algo:main algorithm} [line \ref{algoline:inFunnel check}]) and during sampling too. The check is performed based on \eref{eq:Matrix ellipse equation}, with ellipsoidal inlet regions of funnels projected down to $\config$.
 
\subroutine{compossible}() (\algoref{algo:compossibilty check}) checks whether the funnel-pair is motion-plan compossible as defined in \eqref{def:motion-plan compossibility}. The ellipsoid-in-ellipsoid check in line \ref{algoline:ellipsoid-in-ellipsoid check} is by approximating the outlet-ellipsoid into a convex hull by sampling points on the boundary of the ellipsoid, $\partial\mathcal{E}_o$. The extreme-points are chosen based on singular-value decomposition of the ellipsoid matrix, $M_o$, and checked whether it lies in the interior of $\mathcal{E}_i$ using \eqref{eq:Matrix ellipse equation}. 

\begin{algorithm}
 \caption{\subroutine{computeShortestPathTree}()}
 \begin{algorithmic}[1]
    \State $k_{start} \gets \subroutine{computeKey}(q_{start})$
    \While{$\priorityQ.\subroutine{topKey}() < k_{start} \lor lmc(q_{start}) \neq g(q_{start})$}\label{algoline:compute shortest path while loop}
    \State $v \gets \priorityQ.\subroutine{pop}()$; $k_{old} \gets key(v)$ \label{algoline:repair nodes begin}
        \State $k_{new} \gets \subroutine{computeKey}(v)$
        \If{$k_{new} > k_{old}$} \Comment{check \& update key} \label{algoline:key comparison}
            \State $\priorityQ.\subroutine{push}(v,k_{new})$ 
        \ElsIf{$g(v) > lmc(v)$} \Comment{over-consistent}
            \State $g(v) \gets lmc(v)$
            \State \textbf{for all} {$u \in Pred(v)$} \textbf{do} \subroutine{updateVertex}($u$)
        \Else \Comment{under-consistent}
            \State $g(v) \gets \infty$
            \State \subroutine{updateVertex}($v$)
            \State \textbf{for all} {$u \in Pred(v)$} \textbf{do} \subroutine{updateVertex}($u$)
        \EndIf \label{algoline:repair nodes end}
    \EndWhile
  \end{algorithmic}
  \label{algo:computeShortestPath}
\end{algorithm}

\subsubsection{\textbf{Building the shortest-path subtree} (\algoref{algo:computeShortestPath})}
\label{subsec:computeShortestPath}

Given the search graph, a tree rooted at the goal with minimum cost-to-goal is calculated using techniques outlined in \secref{subsec:discrete graph replanning}. Invoking \subroutine{computeShortestPathTree}() ensures that the robot/start node becomes consistent and also all the nodes with lesser key value than the start node \algolineref{algoline:compute shortest path while loop}. So, in effect the shortest path from each ``promising" node in the search graph to goal, based on cost-to-goal and heuristic values, is quickly determined. 

Inconsistent nodes are popped out of the priority queue, $\priorityQ$ and repaired \algomultilineref{algoline:repair nodes begin}{algoline:repair nodes end}, i.e. made consistent until the robot or start node becomes consistent or the queue becomes empty (usually encountered during the pre-planning phase). The key-comparisons in lines \ref{algoline:compute shortest path while loop} and \ref{algoline:key comparison} are based on lexicographic comparison $-$ the second entry becomes relevant only during tie-breaker among first entries \cite{Koenig.Likachev.AAAI02}. 



\begin{figure*}[t]
    \centering
    \includegraphics[width=0.95\textwidth]{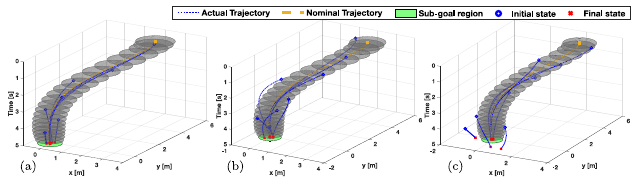}
    \caption{A funnel computed based on Lyapunov level-set theory offers a sufficient but not a necessary condition for invariance. (a) Trajectories starting inside the funnel stay within the funnel; one can not conclusively comment about trajectories starting outside the funnel - it may (b) or may not (c) terminate within the desired goal set (green region)}
    \label{fig:invariance}
\end{figure*}

\begin{algorithm}
 \caption{\subroutine{updateVertex}($v$)}
 \begin{algorithmic}[1]
    \State $lmc(v) \gets \subroutine{computeLMC}(v)$ \label{algoline:computeLMC}
    \State $parent(v) \gets \subroutine{findParent}(v)$ \label{algoline:find parent}
    \If{$v \in \priorityQ$} \label{algoline:remove from queue begin}
    \State $\priorityQ.\subroutine{remove}(v)$ 
    \EndIf \label{algoline:remove from queue end}
    \If{$g(v) \neq lmc(v)$} \Comment{inconsistent} \label{algoline:push to queue begin}
    \State $key(v) \gets \subroutine{computeKey}(v)$
    \State $\priorityQ.\subroutine{push}(v,key(v))$
    \EndIf \label{algoline:push to queue end}
  \end{algorithmic} 
  \label{algo:update vertex}
\end{algorithm}

\subroutine{updateVertex}($v$) (\algoref{algo:update vertex}) computes $lmc$ of vertex $v$ \algolineref{algoline:computeLMC} based on \eref{eq:lmc definition}. The node, $v$ is removed from the priority queue \algomultilineref{algoline:remove from queue begin}{algoline:remove from queue end} and added to the priority queue with the updated key value only if it is inconsistent \algomultilineref{algoline:push to queue begin}{algoline:push to queue end}. The priority value given by \subroutine{computeKey}() is computed using \eref{eq:key for priority queue}. \subroutine{computeHeuristic}($v$) calculates the admissible heuristic value $-$ Euclidean distance from $v$ to $q_{start}$. \subroutine{findParent}($v$) \algolineref{algoline:find parent} determines the best parent of the node, $v$ by analysing its outNeighbors, $N^+(v)$.

\begin{equation}
    parent(v) \leftarrow \argmin_{v' \in N^+} \{c(v,v') + g(v')\}
    \label{eq:determine parent node}
\end{equation}

The priority queue is implemented using a binary heap. The queue operations are briefly described as follows$-$ $\priorityQ.\subroutine{push}(v,key)$ inserts the element, $v$ into the queue at the appropriate place based on key-value. $\priorityQ.\subroutine{pop}()$ removes the top element of the queue and returns it. $\priorityQ.\subroutine{remove}(v)$ removes the entry, $v$ and rebalances the heap. Lastly, $\priorityQ.\subroutine{topKey}()$ returns the key-value of the top-most element in the queue.

\subsubsection{\textbf{Robot motion}}
\label{subsec:robot motion}
The various modules in \algoref{algo:main algorithm}$-$ sensing \algomultilineref{algoline:sensing start}{algoline:sensing end}, planning \algolineref{algoline:replanning} and robot motion \algomultilineref{algoline:motion start}{algoline:motion end} have different operating frequencies. This is implemented by running the methods on separate threads with different frequencies. \subroutine{robotMove}($q_1$, $q_2$) in line \ref{algoline:robot move} determines and applies the corresponding control policy to move from $q_{1}$ to $q_{2}$. The changes to the obstacle set, $\Delta \obstacleSpace$ are estimated using sensors on the robot, and the cost of affected edges are updated using \subroutine{modifyEdgeCosts}($\Delta \obstacleSpace$) (\algoref{algo:modify edge costs - dynamic environment}). The ``head" of the modified edges, i.e. $v$ in $e = (v,w)$ are checked for inconsistencies, and added to the priority queue if inconsistent using \subroutine{updateVertex}() (lines \ref{algoline: update vertex 1} and \ref{algoline: update vertex 2}).

\begin{algorithm}
 \caption{\subroutine{modifyEdgeCosts}($\Delta \obstacleSpace$)}
 \begin{algorithmic}[1]
    \ForAll{$e = (v,w) \in E_m$} \Comment{``motion-edges" set}
        \If{$\neg \subroutine{collisionFree}(e,\Delta \obstacleSpace$)}
            \State $c(v,w) \gets \infty$
            \State \subroutine{updateVertex}($v$) \label{algoline: update vertex 1}
        \Else \Comment{if edges become free}
            \State $c(v,w) \gets c_{prev}$
            \State \subroutine{updateVertex}($v$) \label{algoline: update vertex 2}
        \EndIf
    \EndFor
  \end{algorithmic} 
  \label{algo:modify edge costs - dynamic environment}
\end{algorithm}

\subsubsection{\textbf{Collision checking}}
\label{subsec:collision checking}
We exploit the geometric properties of the funnel and environment to come up with computationally efficient subroutine $-$ \subroutine{collisionFree}($\funnel$, $\obstacleSpace$) for checking overlaps with obstacle set, $\obstacleSpace$. A funnel $\funnel$ is said to be \emph{in-collision} if $\mathcal{P}^\mathcal{S}_\workSpace(\funnel) \cap \obstacleSpace \neq \emptyset$, where $\mathcal{P}^\mathcal{S}_\workSpace(.)$ is the projection of funnel down to the workspace.

Assuming obstacles with locally Lipschitz continuous boundaries, we perform collision-checks between obstacles and the projected level-sets of a funnel. A bounding-volume check constitutes as the first-pass in collision detection. If it fails, the individual ellipsoids of the funnel are checked for collision, in the order given by a Van der Corput sequence \cite{LavalleBook}. We implement a similar method of forming a convex hull around the obstacle and checking whether the extreme points lie within the ellipsoid using \eref{eq:Matrix ellipse equation}. For a general class of obstacles, one can resort to off-the-shelf software such as RoboDK \cite{RoboDK}, MPK \cite{Gipson.MPK} for collision-detection.

\begin{center}
\begin{tabular}{|c|c|}
\hline \textbf{Space} & \textbf{Features $-$ elements, routines, operations} \\
\hline $\mathbb{R}^+$ $\times$ $\stateSpace$ &  funnels, steering, \emph{compossibility-check} \\
\hline \configSpace &  configurations, sampling, \emph{re-planning}\\
\hline Workspace & robot, obstacles, \emph{collision-checking} \\
\hline
\end{tabular}
\end{center}

\subsection{Computing regions of finite-time invariance}
\label{subsec:funnel computation}

Determining a closed-form solution to \eref{eq:Lyapunov condtion} from a general class of Lyapunov functions is not guaranteed, and is computationally intractable. Under certain assumptions such as polynomial closed-loop dynamics, and quadratic Lyapunov candidate functions, the problem of computing the funnels can be reformulated into a Sum-of-Squares (SoS) program \cite{Tedrake.etal.IJRR10}. Consider a quadratic Lyapunov candidate function centred around the nominal trajectory, $\stateVector_0(t)$ defined using a positive definite matrix, $P(t)$
\begin{equation}
    \LyapunovFunction = (\stateVector-\stateVector_0(t))^TP(t)(\stateVector-\stateVector_0(t))
    \label{eq:quadratic Lyapunov function}
\end{equation}

For the class of piecewise polynomials $P(t)$, we solve the SoS program using polynomial $\mathcal{S}$-procedure \cite{Parrilo.MathProgramming03}. The convex optimisation problem of maximising the funnel volume while satisfying constraints \eref{eq:Lyapunov condtion} is solved using bilinear alternation $-$ improving $\rho(t)$ and finding Lagrange multipliers to satisfy negativity of ($\Dot{V}(t,\stateVector) - \Dot{\rho}(t)$) in the semi-algebraic sets \cite{Tobenkin.etal.IFAC11}. This maximises the inner-approximation of the verified regions of invariance around the nominal trajectory \cite{Majumdar.Ahmadi.Tedrake.ICRA13}.

As noted in \cite{Tobenkin.etal.IFAC11}, we observe that time-sampled relaxations in the semi-definite program improve computational efficiency while closely resembling the actual level-sets. Therefore, we leverage this result to carry out optimisations only at discrete time instances between the knot points along the finite time interval.
For $M \in S_+^n$, set of $n \times n$ symmetric, positive definite matrices and $\bm{c} \in \mathbb{R}^n$, \eref{eq:Matrix ellipse equation} represents an ellipsoid centred around $\bm{c}$.
\begin{equation}
    (\bm{x}-\bm{c})^TM(\bm{x}-\bm{c}) = 1
    \label{eq:Matrix ellipse equation}
\end{equation}

The invariant sets, $\levelSet(t)$ in \eref{eq:Lyapunov level set} corresponding to the quadratic Lyapunov function defined in \eref{eq:quadratic Lyapunov function} are the closed set, i.e. interior and boundary of ellipsoids centered around the nominal trajectory, $\stateVector_0(t)$.

\begin{equation}
    \mathcal{E}(t) = P(t)/\rho(t)
    \label{eq:ellipsoids}
\end{equation}

The closed-loop dynamics is derived using the feedback control policy, $\bm{u}(t,\stateVector)$ and the system equations. Given a nominal trajectory in finite time interval, the dynamics are approximated to polynomial equations about it using Taylor-series expansion. The region of desired final-states, referred to as \emph{sub-goal}, $\mathcal{X}_f$ is assumed to be an ellipsoid, defined by $\mathcal{E}_f$ centred at the final state. Computing the \emph{maximal inner-approximation} of the \emph{backward-reachable invariant set} to the sub-goal region is formulated as an SOS program, and the resulting semi-definite program (SDP) is numerically solved.

Sample funnels computed for quadrotor dynamics with nominal control (presented in \secref{subsec:dynamics}) have been illustrated in \figref{fig:invariance}. Note that the funnels have been projected from $\{t\} \times \stateSpace$ down to lower dimensional subspace, $\{t\} \times \mathbb{R}^2$ for visualisation. Funnels calculated using above-mentioned methods have formal guarantees of invariance, ensuring that robot-trajectory stays within the backward-reachable set, if it starts within the funnel (see \figref{fig:invariance}-a).

\begin{figure}[b!]
    \centering
    \includegraphics[width=0.7\columnwidth]{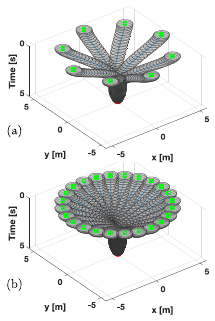}
    \caption{Funnel library, $\mathfrak{L}$. The initial configurations at $t=0$ lie at an $\epsilon$-distance from $\bm{0}$ ($\epsilon = 5m$). Desired final-states is a compact set centered around origin with radius $0.3m$. Using $\mathfrak{L}_b$ (bottom) will result in a finer resolution of motion-plan than $\mathfrak{L}_a$ (top)}
    \label{fig:funnel library}
\end{figure}

\subsection{Designing the funnel library}
\label{subsec:funnel library design}

We deal with Lagrangian systems with time-invariant dynamics a.k.a autonomous systems. Considering the state-feedback controller, the closed-loop dynamics reduces down to $\Dot{\stateVector}(t) = \bm{f}(\stateVector(t))$. Hence, for ease of usage we shift the initial time to origin, $t_0 = 0$, and the time interval becomes $[0, T_f]$, where $T_f \in R^+$ is a finite non-negative real number. 


The funnel library, $\funnelLibrary$, consists of a finite number of verified trajectories, encapsulating the information of the certified regions of invariance in finite-time interval. Each funnel, $\funnel_i \in \funnelLibrary$, is parametrized by the nominal trajectory, $\stateVector_{0_i}(t)$, the ellipsoidal level-sets, $\mathcal{E}_i(t)$ and the final time, $T_f$. The trajectories and the ellipsoids are projected from the state-space onto $\config$-space using an appropriate projection operator, $\mathcal{P}^\mathcal{S}_\config(.): \stateSpace \rightarrow \config$. The resultant projection of invariant sets take the form of ellipsoids when projected onto the $d$-dimensional Euclidean subspace of robot-configurations \cite{Karl.Ellipsoid.1994},

\begin{equation}
    S(t) = (B^T\mathcal{E}(t)^{-1}B)^{-1}
    \label{eq:ellipse projection}
\end{equation}
where $B$ is a $n \times d$ matrix, consisting of the basis vectors of the coordinates of the $\config$-space in state space. Additionally, the funnels in the library are projected down to the robot-workspace in the pre-processing phase to speedup collision-checking with obstacles during runtime.

The funnel library acts like a bridge between the offline and online phases $-$ invariant set analysis and motion planning. Therefore, certain algorithm parameters such as extend-distance, resolution of the planner, range of obstacle sizes, etc. is considered while constructing the library. Meanwhile, the funnel library provides the vital information of compossibility required during the online phase of motion planning.

\figref{fig:funnel library} illustrates examples of funnel libraries, with nominal trajectories starting at an $\epsilon$-distance from origin, at various translational positions and terminating in a sub-goal region centered around origin at final time, $T_f$. These invariant sets can be shifted along the cyclic coordinates, and time to ensure the final pose of the system state aligns with the sampled configurations. \figref{fig:funnel library}-a depicts a sparser library  which would result in lesser resolution of the motion plan.

It is worth mentioning that the initial conditions of the finite number of projected trajectories in the library, along with the appropriate shift operator, $\Psi_c(.)$ along the cyclic coordinates should be able to span the entire $\config$-space. This ensures \emph{probabilistic coverage} \cite{Tedrake.etal.IJRR10} of the sampling-based motion re-planning algorithm and hence, probabilistic completeness.


\section{EXPERIMENTAL VALIDATION}

\label{sec:validation}
We validate our algorithm on a quadrotor UAV in simulations, flying through an indoor space with dynamic obstacles. This section briefly describes the dynamics of a quadrotor with a nominal controller, and discusses our experiments with implementation details. We demonstrate the relevance of invariant sets and empirically guarantee completeness and correctness of our motion re-planning algorithm.

\subsection{System Dynamics \& Mission Profile}
\label{subsec:dynamics}
The equations of motion of a quadrotor UAV are derived using Newton-Euler formulation \cite{Garcia.etal.Springer06}. Considering position, $\bm{\xi} = [x, y, z]^T$ and the attitude, $\bm{\eta} = [\phi, \theta, \psi]^T$ of the quadcopter defined in an inertial frame, the equations can be written as,
\begin{align}
\begin{split}
\label{eq:quad dynamics}
&\dot{\bm{\xi}} = \bm{v} \\
&m\dot{\bm{v}} = -mg\bm{e_3} + R\bm{e_3}T \\
&\dot{\bm{\eta}} = W_\eta\bm{\omega} \\
&J\dot{\bm{\omega}} = -\bm{\omega} \times J \bm{\omega} - J_r(\bm{\omega} \times \bm{e_3})\Omega + \bm{M}
\end{split}
\end{align}
Where, $m$ is the mass, $J$ the inertia matrix, $\bm{v}$ the linear velocity and $\bm{\omega}$ is body angular rates. $\bm{e_3}$ is $\begin{bmatrix} 0 & 0 & 1 \end{bmatrix}^T$, $J_r$ is inertia of the rotor, $\Omega = \Omega_1 - \Omega_2 + \Omega_3 - \Omega_4$ is the net rotor speed. $\Omega_i$ denotes the rotational speed of the individual rotors.

\begin{figure}[t]
    \centering
    \includegraphics[width=\columnwidth]{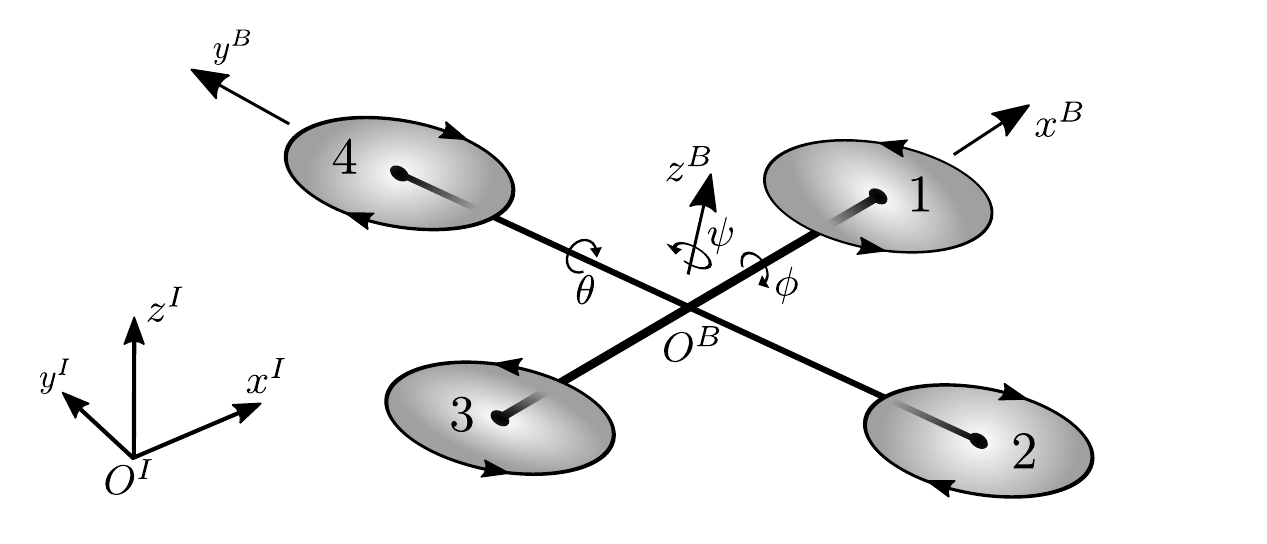}
    \caption{Schematic of a quadrotor with 6DOF (position and attitude) in inertial, $I$ and body-fixed, $B$ frames of reference}
    \label{fig:quad schematic}
\end{figure}

$R \in \mathcal{SO}(3)$ is the rotation matrix from the body frame to the inertial frame. $W_\eta$ is the transformation matrix for angular velocities in the body frame to inertial frame. For the specific configuration of rotors as in \figref{fig:quad schematic}, Thrust, $T$ and Moment, $\bm{M}$ are defined as,
\begin{align}
\label{eq:mixer model}
\begin{split}
T &= k\sum_{i=1}^4 \Omega_i^2 \\
\bm{M} = \begin{bmatrix} M_\phi \\ M_\theta \\ M_\psi \end{bmatrix} &= \begin{bmatrix} kl(\Omega_4^2 - \Omega_2^2) \\ kl (\Omega_3^2 - \Omega_1^2) \\ d (\Omega_2^2 + \Omega_4^2 - \Omega_1^2 - \Omega_3^2) \end{bmatrix}
\end{split}
\end{align}
where  $k$ is the thrust coefficient, $d$ the counter-moment drag coefficient and $l$ is the arm length.

The state of the system, $\bm{x} = [\bm{\xi} \tab[0.12cm] {\bm{v}} \tab[0.12cm] {\bm{\eta}} \tab[0.12cm] \bm{\omega}]^T$ with $4$ rotor speeds as inputs, $\bm{u} = [\Omega_1 \tab[0.12cm] \Omega_2 \tab[0.12cm] \Omega_3 \tab[0.12cm] \Omega_4]^T$. Typically, the controller architecture has a cascaded structure, with a fast inner loop stabilising the attitude and a outer loop tracking the position or velocities  \cite{Jaffar.etal.ICC19}. We implement a nested P-PID loop for attitude-tracking. Based on the desired angles, the proportional controller computes the desired angular body rates which are then tracked using a PID controller. This has been found to be effective in maneuvers which don't require large deviations from nominal hover conditions \cite{Luukkonen.Report11}. The outer loop tracks the desired position setpoints and is achieved using an LQR controller \cite{Bouabdallah.IROS04}. Equivalently, the inputs to the quadrotor position controller are the desired setpoints $-$ $[x_d, y_d , z_d, \psi_d=0]^T$.

The configuration space, $\config = \mathbb{R}^3$, whereas the state-space of the quadrotor translational subsystem is $\mathcal{S} = \mathbb{R}^3 \times \mathbb{R}^3$, comprising of the position and velocities. The mission profile is to fly at a set altitude, $z_d = h$ with a zero heading-angle, $\psi_d = 0$. Owing to the reduced operation-space, the workspace and the sampling is in $\mathbb{R}^2$, and it suffices to check for possible collisions in the 2D plane between ellipses and obstacles. The start configuration and goal region are defined in the $xy$-plane. The library is appropriately constructed, see \figref{fig:funnel library}. 

\begin{figure*}[t]
    \centering
    \includegraphics[width=\textwidth]{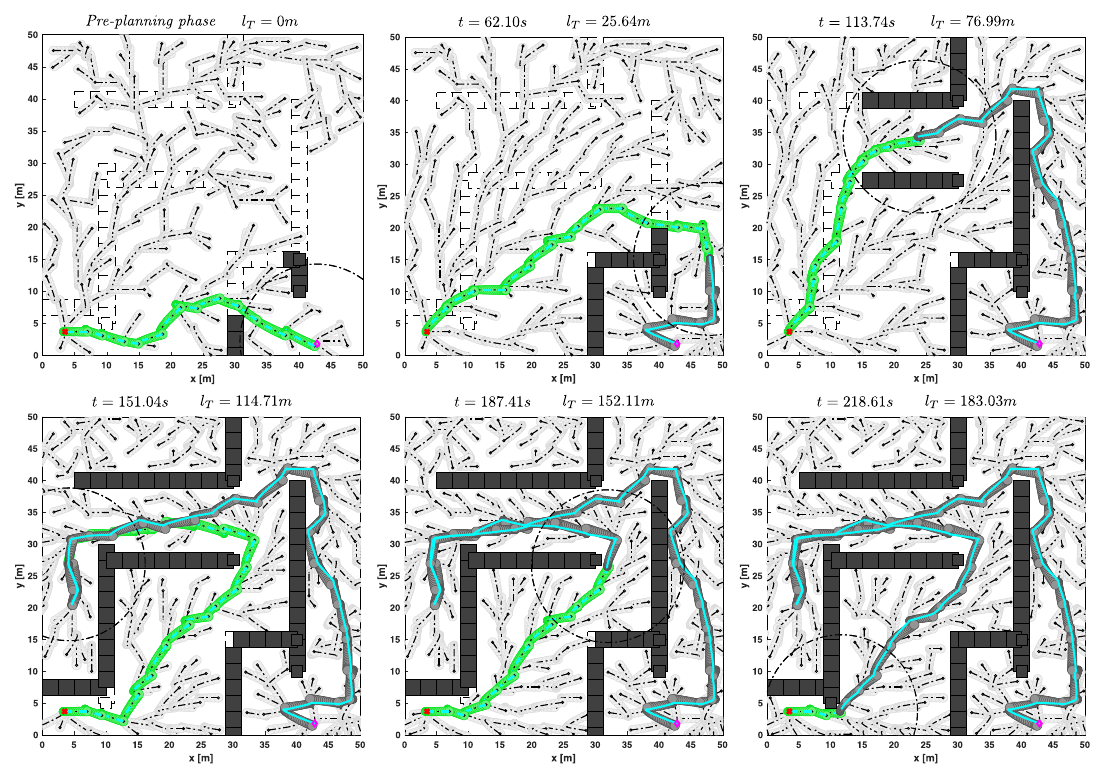}
    \caption{Time instances of motion plan executed by PiP-X on a quadrotor flying with altitude-hold. The start configuration is in the lower-right corner, with the goal location at lower-left corner. The quadrotor senses obstacle-walls (solid rectangles) within sensor-radius (dashed circle), and recomputes motion plans (green \emph{funnel-path}) accordingly. The traversed funnels and funnel search-tree are denoted by dark and light gray, respectively. $l_T$ denotes the traversed-path length, and $t$ denotes time elapsed. Note that the robot-trajectory (cyan) with the funnels and obstacles are projected down to $(x,y)$ subspace}
    \label{fig:results - motion re-plan time splices}
\end{figure*}

\subsection{Experimental setup}

The equivalent \emph{closed-loop position} dynamics of the quadrotor is derived from repeated trials with various position setpoints, $\bm{\xi}_d$ given as inputs to the system. 
\begin{equation}
    \dot{\bm{\xi}} = \bm{v} \tab[0.8cm] \dot{\bm{v}} = \bm{f}(\bm{\xi},\bm{v},\bm{u}) \tab[0.8cm] \bm{u} = \bm{\xi}_d
\end{equation}

The system identification of the translational subsystem, $\bm{x} = [\bm{\xi} \tab[0.2cm] \bm{v}]$, is carried out using SysId toolbox in MATLAB. The identified equations are then approximated to polynomial dynamics using a third-order Taylor-series expansion about the nominal trajectory. An estimate of the required final time of finite-interval, $[0,T]$ is obtained based on the time taken by the system to reach within the defined goal region of $0.3m$ around a desired setpoint. Subsequently, the invariant sets centered around the nominal trajectory are constructed using the methods described in \secref{subsec:funnel computation}. The Sum-of-Squares optimisation is converted to an SDP by Systems Polynomial Optimisation Toolbox (SPOT) in MATLAB and solved using SeDuMi \cite{Sturm.Sedumi}.

Our reverse-search algorithm requires backward-reachable set $-$ starting from different initial conditions, the desired setpoint is given as the origin, $\bm{\xi_d} = \bm{0}$. The initial $(x,y)$ conditions lie on an $\epsilon$-circle centered at origin, with $z(0) = h$ and $\psi(0) = 0$, as shown in \figref{fig:funnel library}. All the funnels computed are stored in a dictionary with a key-identifier based on the starting location. Each funnel in the library is characterised by its key, the nominal trajectory, ellipsoidal invariant sets and nominal control inputs - rotor speeds, $\Omega_i$ at discrete time-steps within the finite-time horizon. 

In order to verify the motion plan, we develop a higher fidelity model in Simulink. The dynamics are based on \eref{eq:quad dynamics} - \eqref{eq:mixer model}. In addition to that, we incorporate actuator saturations, rotor dynamics and process noise. We believe that these additions will enable the simulation model to closely resemble the physical system. Using the funnel library as in \figref{fig:funnel library}-b, our algorithms are tested in two different 2D environments, \emph{random forest} and \emph{maze}, with various types of obstacle-changes, described in the next section. 

\subsection{Results}

\begin{figure*}[t]
    \centering
    \includegraphics[width=0.87\textwidth]{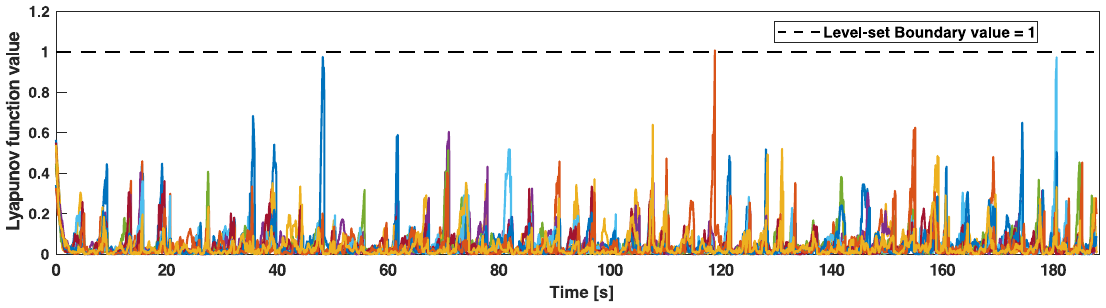}
    \caption{\emph{Normalised} Lyapunov function value of the system's state simulated using the higher-fidelity quadrotor model, with solution funnel-path given by the algorithm across $10$ different trials (denoted by different line-colors)}
    \label{fig:results - lyapunov function value}
\end{figure*}

A sample run of PiP-X in a user-specified maze environment is shown in \figref{fig:results - motion re-plan time splices}. The robot perceives obstacle-walls of the maze within a finite sensing radius and updates its motion plans accordingly.  We notice that a considerable amount of the $\config$-space is explored with fewer samples. Hence, using volumetric verified trajectories potentially speeds up the rate of probabilistic coverage. The position setpoints in the funnel-path to goal, output by the algorithm are given to the simulation system in real-time, and the system's actual trajectory is analysed. It is observed that the trajectory of the system lies within the solution funnel-path throughout the course of the mission profile, verifying set-invariance. 

In another scenario within the same maze environment with different start/goal location, we analyse the normalised Lyapunov function value of the system, \eref{eq:quadratic Lyapunov function} - \eref{eq:ellipsoids}. From \figref{fig:results - lyapunov function value}, we notice that the trajectory stays within the level-set boundary of $V = 1$ till the quadrotor-system reaches the goal region, empirically proving invariance. The peaks in the Lyapunov-function value mostly occur in the outlet/inlet region between 2 funnels. 

Our algorithm, based on quick graph-based replanning, is able to repair motion plans on-the-fly, ensuring a sequence of safe trajectories that are dynamically feasible. The theoretical guarantee of set-invariance enables our algorithm to rewire motion plans, implicitly addressing the two-point BVP encountered during search-tree rewiring in most sampling-based motion planners. Computing a shortest-path tree rooted at the goal results in an optimal path with respect to the iteratively-constructed underlying search-graph. 

We extensively test our algorithm in two different kinds of environments $-$ initially unknown, and dynamically changing $-$ random forest and maze, across various scenarios and conditions; considering algorithm success and length of traversed trajectory as performance metrics. Algorithm failure is defined to be the robot's inability to compute a feasible motion plan within a user-defined timeout or its collision with an obstacle.

\subsubsection{\textbf{Initially unknown Random Forest with robot-sensing}} 

We consider a 2D workspace of dimensions $50m \times 50m$ with circular obstacles of random sizes within the range of [$2$ $4$]$m$, and at random locations. The sensing radius of the quadrotor is $12m$. Each scenario is characterised by the number of tree-obstacles, $N_{t}$ present in the environment. In each scenario, we vary the start and goal configurations, and obstacle locations. The start and goal locations are spaced out $40m$ diametrically apart, for uniformity while comparing performance. The goal region is defined to be a $\delta$-ball of radius $0.3m$ centered at the goal configuration.

\begin{figure}[t]
    \centering
    \includegraphics[width=\columnwidth]{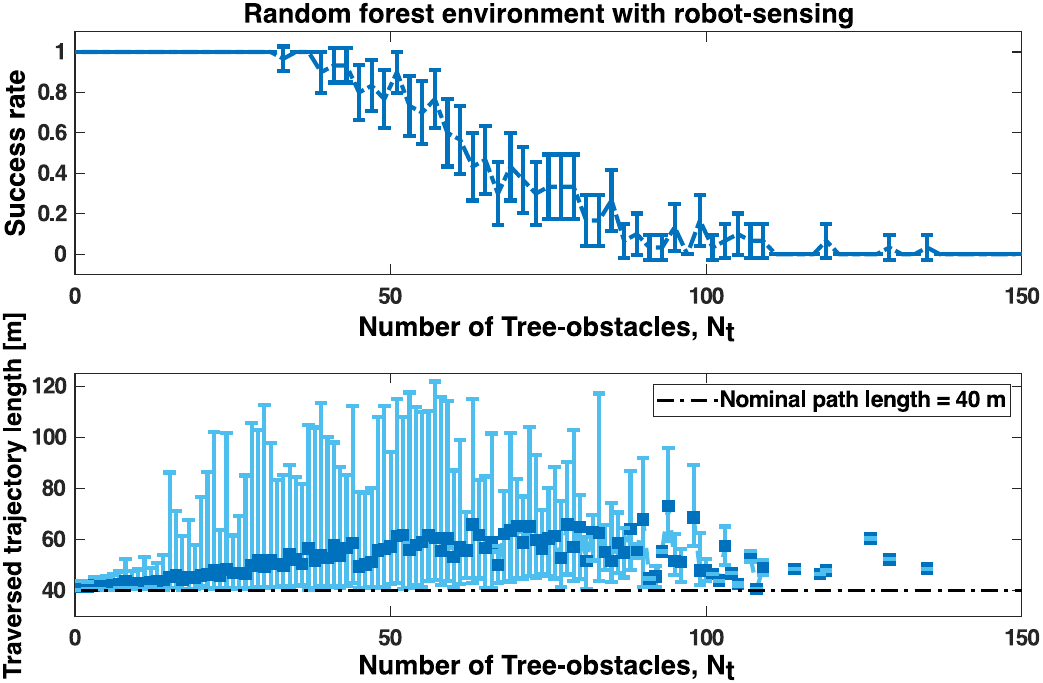}
    \caption{Performance metrics in random forest environment with finite-sensing (a) mean and standard deviation of success rate (b) mean and range of traversed-trajectory length, across $30$ trials}
    \label{fig:results - forest sensing}
\end{figure}

The number of tree-obstacles are varied from $0$$-$$150$ in increments of $1$. $30$ different trials are run in each scenario, reporting mean/standard deviation of success rate, and mean/range of traversed-trajectory length (see \figref{fig:results - forest sensing}). The nominal path length, i.e if there are no obstacles and the robot is holonomic, is $40m$.

We observe that the algorithm has $100\%$ success rate until $N_t = 32$ ($14.4\%$ of workspace being obstacle space). The success rate drops from $1$ to $0$ as $N_t$ increases. The algorithm fails after $134$ trees in the forest, which approximately translates to $61.04\%$ of workspace covered with obstacles. As expected, path length increases with number of obstacles, until we start encountering algorithm failure; wherein a few successful trials skews the average trajectory-length.

\subsubsection{\textbf{Random Forest with dynamic obstacles}}

Similar to the previous workspace, this environment is such that tree-obstacles are deleted and added at random, emulating a dynamic setup. The changes occur anywhere in the workspace and the robot is capable of sensing all those changes. A scenario is described by number of trees, $N_t$ and change-percentage, $C$. For e.g., a change, $C = 60\%$ in a workspace with $N_t = 45$ implies $27$ pre-existing trees are removed and 27 new obstacles are added $-$ changing location and size. $C = 0\%$ trivially refers to a static environment.

\begin{figure}[t]
    \centering
    \includegraphics[width=0.76\columnwidth]{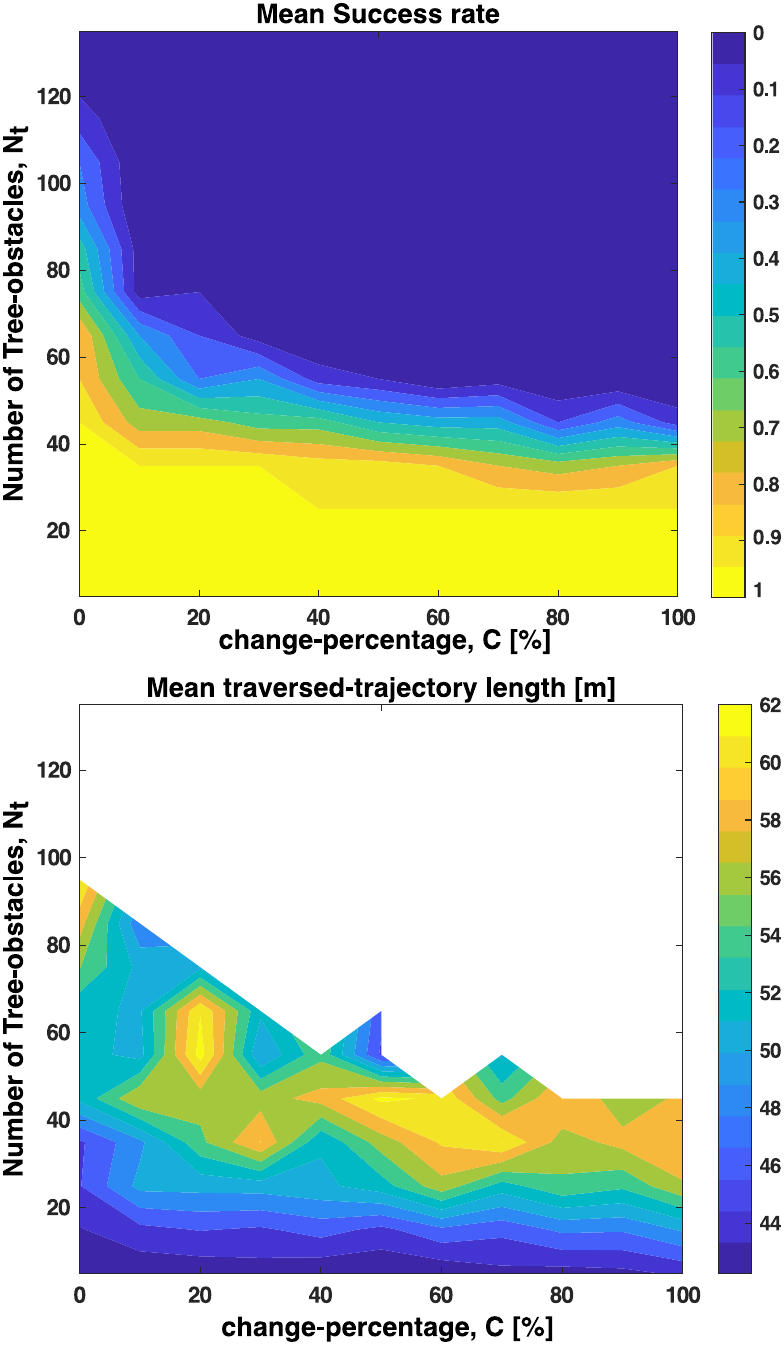}
    \caption{Forest environment with dynamic obstacles $-$ mean of success-rate and traversed-trajectory length from $25$ trials}
    \label{fig:results - forest dynamic}
\end{figure}

Taking input from previous experimental analysis, we consider the range of $N_t$ to be [$5$, $135$] in increments of $10$. The change-percentage is varied from $0$ to $100$ in steps of $10$. We run $25$ trials (different start/goal configurations) and report mean of success rate and traversed-trajectory length in the form of a contour plot$-$ \figref{fig:results - forest dynamic}.

We observe that the algorithm failure and mean trajectory-length increases with either increasing number of tree-obstacles, $N_t \geq 45$, or higher level of changes, $C \geq 50$. It completely fails when the environment is densely cluttered with obstacles or highly dynamic (upper right triangle of the contour in \figref{fig:results - forest dynamic}-a). Note that trajectory-length in static environments ($C = 0\%$) with $N_t \geq 95$ is not visualised in \figref{fig:results - forest dynamic}-b as they are isolated instances of algorithm successes.

\subsubsection{\textbf{Initially-unknown Maze with finite robot-sensing}}

A maze-like environment with rectangular walls is designed in a two-dimensional $50m \times 50m$ workspace, similar to the one in \figref{fig:results - motion re-plan time splices}. The robot senses the obstacle-walls within a limited radius of $12m$.

\begin{figure}[t]
    \centering
    \includegraphics[width=0.9\columnwidth]{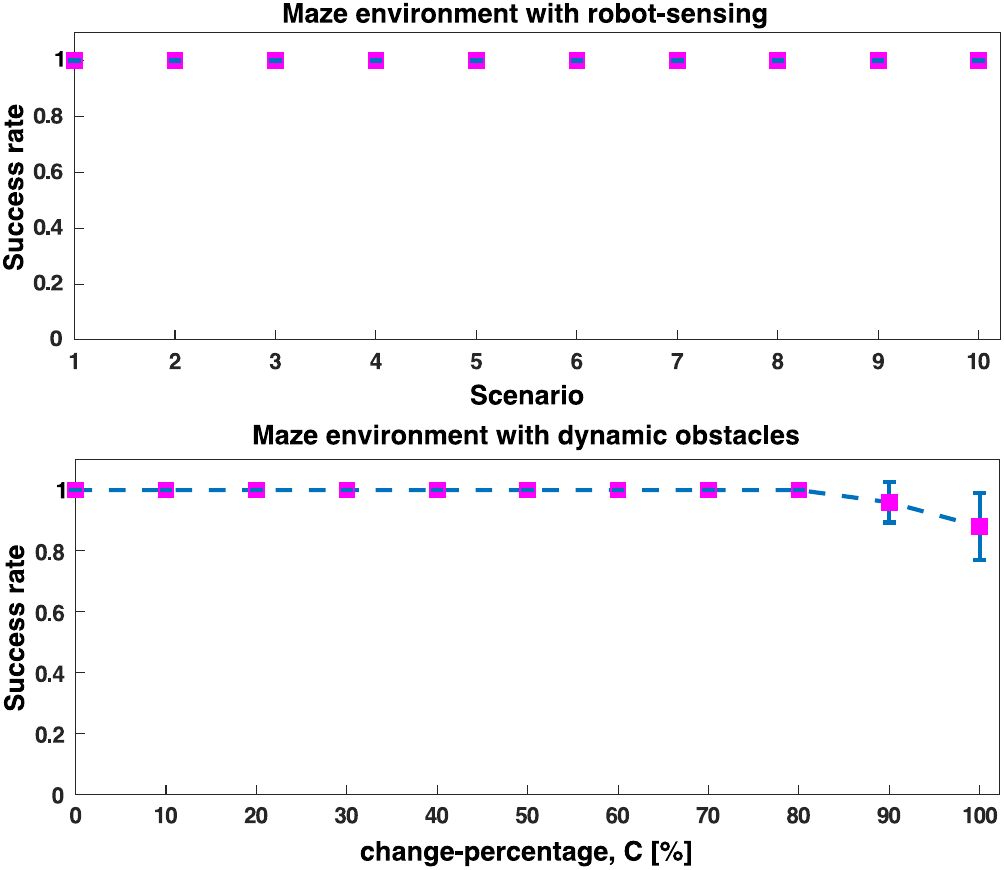}
    \caption{Mean and standard deviation of algorithm success rate from $25$ trials in a maze-environment with (a) finite robot-sensing, and (b) dynamic obstacles}
    \label{fig:results - maze}
\end{figure}

$25$ trials are run for $10$ different scenarios of start/goal configurations, and mean/standard deviation of success rate is presented in \figref{fig:results - maze}-a. Since the start and goal locations are such that a solution path exists, our algorithm is always capable of computing an initial motion plan and accordingly replan as new obstacle-walls are perceived. This results in a $100\%$ success rate across $10$ different scenarios.

\subsubsection{\textbf{Maze with dynamic changes}}

Similar to the previous environment, we additionally incorporate dynamic obstacles in the form of ``windows" that can open (obstacle-deletion) or close (obstacle-addition) at random, and the robot is capable of sensing such obstacle-changes. 

The maze has $20$ walls and $10$ windows $-$ $33.8\%$ of wall-length (assuming uniform obstacle-width). Change-percentage, $C$ refers to how many of the windows open and close at sensing-frequency $-$ it is varied from $0\%$ (static) to $100\%$ (dynamic) in increments of $10$. From \figref{fig:results - maze}, we observe a drop in success at $C = 90\%$ and $C = 100\%$ (highly dynamic). Note that we consider only success rate because the path length depends on the contour of the maze and start/goal configurations $-$ there is no nominal path length for reference.

As a general observation from all experiments and scenarios, most failures are due to ``idleness" time-outs, $I_M$ implying the algorithm's inability to identify and report that a solution doesn't exist, as is with the case of all sampling-based motion planning techniques. Another common reason for failure is the algorithm's inability to fit a volumetric region of space in narrow gaps, especially in dense-cluttered environments. In scenarios with highly-dynamic obstacles, an obstacle is more probable to appear on the traversing funnel-edge, inevitably leading to a collision with the obstacle. 


  


\section{CONCLUSIONS}
\label{sec:conclusion}

A novel sampling-based online feedback motion re-planning algorithm using funnels, PiP-X is presented. The search graph with funnel-edges is iteratively constructed using sampling-techniques, and concurrent calculation of the shortest-path subtree of funnels rooted at the goal ensures optimal funnel-path from robot configuration to the goal region. The use of incremental graph-replanning techniques and a pre-computed library of motion primitives ensure that our method can repair paths \emph{on-the-fly} in dynamic environments.

The information of robot-traversability and funnel-sequencibility is represented together in the form of an \emph{augmented directed-graph}, helping us leverage numerical graph-search methods to compute safe, controllable motion-plans. Analysing and formally quantifying stability of trajectories using Lyapunov level-set theory ensures kinodynamic feasibility of the solution-paths. Additionally, verifying the compossibility of a funnel-pair proves to be a ``relaxed" alternative to the \emph{two-point boundary value problem}, encountered in most single-query sampling-based motion planners that require rewiring.

Our technique is validated on a simulated quadrotor platform in a variety of environments and scenarios. Performance of the algorithm in terms of success and trajectory-length is examined. Our approach combines concepts from systems analysis, sampling-based methods, and incremental graph-search to address \emph{feedback motion re-planning} for any general nonlinear robot-system in dynamic workspaces.

\section*{ACKNOWLEDGMENTS}
The authors are grateful to the Robot Locomotion Group at MIT for providing an open-source software distribution for computing funnels. We like to thank Sharan Nayak for the valuable discussions on this work. We also thank the members of Motion and Teaming Lab, UMD $-$ Loy McGuire, Alex Mendelsohn, Alkesh K. Srivastava and Dalan C. Loudermilk for their feedback during the preparation of this manuscript. 


\balance

\bibliographystyle{IEEEtran} 

\begin{IEEEbiography}[{\includegraphics[width=1in,height=1.25in,clip,keepaspectratio]{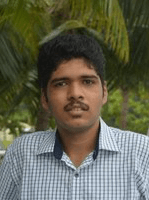}}]{Mohamed Khalid M. Jaffar} received his Dual Degree, Bachelor of Technology (B.Tech.) and Master of Technology (M.Tech.) in Aerospace Engineering from the Indian Institute of Technology Madras, Chennai, India in 2018. He is currently pursuing a Ph.D. degree in Aerospace Engineering, under the supervision of Dr. Michael Otte.
He is currently a Research Assistant with the department of Aerospace Engineering, University of Maryland, College Park, MD, USA. His research focuses on the use of control and systems theory for motion planning of aerial robots. 
\end{IEEEbiography}

\begin{IEEEbiography}[{\includegraphics[width=1in,height=1.25in,clip,keepaspectratio]{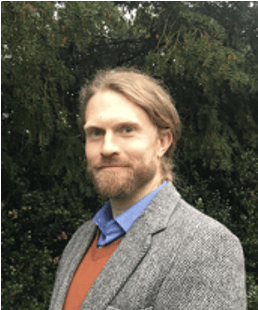}}]{Michael Otte}
(M’07) received the B.S. degrees in aeronautical engineering and computer science from Clarkson University, Potsdam, New York, USA, in 2005, and the M.S. and Ph.D. degrees in computer science from the University of Colorado Boulder, Boulder, CO, USA, in 2007 and 2011, respectively.
From 2011 to 2014, he was a Postdoctoral Associate at the Massachusetts
Institute of Technology. From 2014 to 2015, he was a Visiting Scholar at
the U.S. Air Force Research Lab. From 2016 to 2018, he was a National
Research Council RAP Postdoctoral Associate at the U.S. Naval Research
Lab. He has been with the Department of Aerospace Engineering, at the
University of Maryland, College Park, MD, USA, since 2018. He is the
author of over 30 articles, and his research interests include autonomous robotics, motion planning, and multi-agent systems.
\end{IEEEbiography}



\end{document}